\documentclass[10pt,twocolumn,letterpaper]{article}

\usepackage[pagenumbers]{cvpr} 

\usepackage{graphicx}
\usepackage{amsmath}
\usepackage{amssymb}
\usepackage{booktabs}

\usepackage[pagebackref,breaklinks,colorlinks]{hyperref}

\newcommand{\PAR}[1]{\vspace{0.1cm}\noindent{\bf #1} }

\usepackage[capitalize]{cleveref}
\crefname{section}{Sec.}{Secs.}
\Crefname{section}{Section}{Sections}
\Crefname{table}{Table}{Tables}
\crefname{table}{Tab.}{Tabs.}

\begin{document}

\title{BALF: Simple and Efficient Blur Aware Local Feature Detector}

\author{Zhenjun Zhao\textsuperscript{1,2} \qquad Yu Zhai\textsuperscript{1} \qquad Ben M. Chen\textsuperscript{1} \qquad Peidong Liu\textsuperscript{3,4}\thanks{Corresponding author.}\\
\textsuperscript{1}The Chinese University of Hong Kong \qquad \textsuperscript{2}Peng Cheng Laboratory \\ \textsuperscript{3}Westlake University \qquad \textsuperscript{4}Westlake Institute for Advanced Study\\
{\tt\small zjzhao@mae.cuhk.edu.hk} \quad {\tt\small zhaiyu@link.cuhk.edu.hk} \quad {\tt\small bmchen@cuhk.edu.hk} \quad {\tt\small liupeidong@westlake.edu.cn}
}
\maketitle

\begin{abstract}
   Local feature detection is a key ingredient of many image processing and computer vision applications, such as visual odometry and localization. Most existing algorithms focus on feature detection from a sharp image. They would thus have degraded performance once the image is blurred, which could happen easily under low-lighting conditions. To address this issue, we propose a simple yet both efficient and effective keypoint detection method that is able to accurately localize the salient keypoints in a blurred image. Our method takes advantages of a novel multi-layer perceptron (MLP) based architecture that significantly improve the detection repeatability for a blurred image. The network is also light-weight and able to run in real-time, which enables its deployment for time-constrained applications. Extensive experimental results demonstrate that our detector is able to improve the detection repeatability with blurred images, while keeping comparable performance as existing state-of-the-art detectors for sharp images. 
\end{abstract}

\section{Introduction}
\label{sec:intro}
\begin{figure}[htb]
	\centering
	\includegraphics[width=0.9\linewidth]{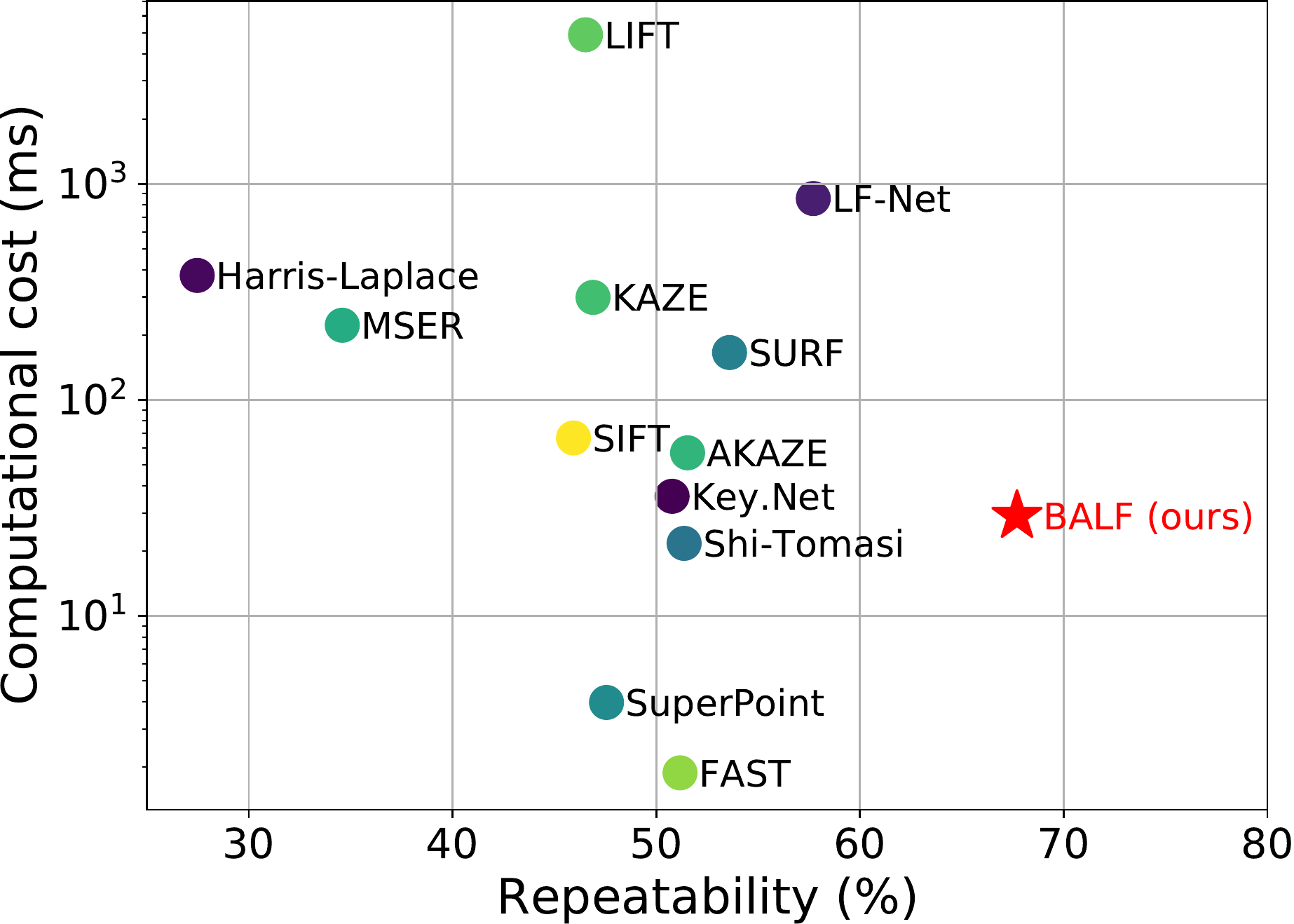}
	\caption{\textbf{The performance of keypoint detectors with motion blurred images.} Our approach achieve superior detection performance in terms of the repeatability metric and efficiency, which paves the way for robust 3D vision under low-lighting conditions.}
	\label{fig_teaser}
	\vspace{-0.5em}
\end{figure}

Being able to accurately detect and describe salient keypoints across images is crucial in many applications such as Simultaneous Localization and Mapping (SLAM), Structure-from-Motion (SfM), camera calibration, video compression, tracking, image retrieval, and visual localization. Interest points should be sparse, repeatable, and discriminable, in order to be extracted and matched across images from different lighting conditions or viewpoints. While mangy state-of-the-art methods have been proposed, motion blur is still a major challenge remaining for local feature detection methods. Motion blur is one of the most common artifacts that degrade images. It usually occurs in low-light conditions where longer exposure times are necessary. This would affect many feature based approaches, which struggle to detect repeatable keypoints to build up correspondences.

To detect keypoints in blurred images, a straightforward way is to utilize a deblurring algorithm to restore the latent sharp image and then to detect keypoints from the restored image. Image/video deblurring methods have been well developed over the last decades, which mainly consist of classic gradient-descent methods \cite{Bahat2017NonuniformBD, Xu2010TwoPhaseKE, Goldstein2012BlurKernelEF, Pan2014DeblurringTI, Perrone2014TotalVB, Pan2016BlindID, Cho2009FastMD, Xu2013UnnaturalLS, Ren2018DeepND, Krishnan2011BlindDU} and learning-based methods \cite{Schuler2016LearningTD, Sun2015LearningAC, Xiao2016LearningHF, Nah2017DeepMC, Tao2018ScaleRecurrentNF, Kupyn2019DeblurGANv2D, Kupyn2018DeblurGANBM, Xu2014DeepCN, Zhang2018DynamicSD, Xu2018MotionBK, Li2018LearningAD, Hradi2015ConvolutionalNN, Liu2020SelfSupervisedLM}. Although deblurring algorithms have achieved impressive performance recently, there are still several limitations existed. For example, existing state-of-the-art methods usually require high computational resources and are hardly to run in real-time even with a high-end GPU. Another limitation is that current methods still cannot perform very well and might introduce additional artifacts for severe motion blurred image, due to the limited information preserved by a single blurred image. We thus aim to design a novel one-stage efficient local feature detector from a motion blurred image directly, without any intermediate deblurring operation, to avoid those drawbacks.


In recent years, deep learning techniques have shown great success in improving local feature extraction. Many state-of-the-art methods have been proposed by the community \cite{Verdie2015TILDEAT, Savinov2017QuadNetworksUL, Zhang2017LearningDA, Laguna2019KeyNetKD, Benbihi2019ELFEL, Tian2020D2DKE, Suwanwimolkul2021LearningOL}. Even though great progress has been achieved in extracting local features, most of them focus on improving the robustness against viewpoint changes, illumination changes etc. Prior work that aims to extract features from motion blurred images is still limited, which is important towards robust 3D vision at low-lighting scenarios (e.g. augmented reality for outdoors at night).
%
%
Further inspired by the recent success of Multi-Layer-Perceptron (MLP) in many areas of computer vision \cite{Tolstikhin2021MLPMixerAA, Liu2021PayAT, Yu2021S2MLPv2IS, Tu2022MAXIMMM, Lian2022ASMLPAA, Chen2022CycleMLPAM, Choe2022PointMixerMF, Mansour2022ImagetoImageMF, Tang2022AnIP, Touvron2022ResMLPFN, Guo2022HireMLPVM, Li2022BraininspiredMP, Hou2022VisionPA}, we propose to explore the possibility by applying MLP-based network for local feature detection from a motion blurred image, which has never been attempted previously.

In this paper, we introduce {\textit {BALF}}, a simple yet both efficient and effective motion blur aware local feature detector. Our detector network consists of two main components: a pure MLP-based image encoder and a keypoint detection module. The encoder takes advantage of a cascaded of multi-axis gated MLP blocks and ``Squeeze and Excitation (SE)'' MLP blocks, to learn a pyramid feature representation of the input image. The keypoint detection module then apply a differentiable channel-wise softmax operator to detect salient keypoints. 
%
%
Extensive evaluations have been conducted on both synthetic and real datasets. Experimental results demonstrate that our method achieves superior performance over prior methods as shown in \cref{fig_teaser}. In particular, we achieve 15\% improvements over the current best performing network on motion blurred images with the repeatability metric, which is commonly used for local feature detection evaluations. Besides the repeatability performance, our method is also able to run at around 35 FPS for VGA resolution image (\ie 480$\times$640 pixels) on a consumer-grade Graphic Card (\ie NVIDIA Geforce 2080 Ti), which is sufficient for time-constrained applications.

In summary, our {\bf{contributions}} are as follows:
\vspace{-0.4em}
\begin{itemize}
	\itemsep0em
	\item We propose a novel and efficient MLP-based network architecture for local feature detection, which has never been attempted for this task previously. 
	\item Extensive experimental results demonstrate that our network achieves superior detection performance over prior works on motion blurred images, while keeping comparable performance for sharp images. 
	\item Our motion blur robust keypoint detector is able to run in real-time, which would enable many time-constrained applications (\eg robotic navigation in low-lighting scenarios).
\end{itemize}

\section{Related Work}
\label{sec:related_work}
We review three main areas of related work: keypoint detection, image deblurring, and MLP related methods. 

\PAR{Keypoint detection.}
Local feature detection plays a vital role in many vision related tasks, such as visual localization and recognition. It thus received a continuous influx of attention in the past decades \cite{Csurka2018FromHT, Gauglitz2011EvaluationOI, Salahat2017RecentAI}. Existing works can be generally categorized into classical handcrafted based methods and modern learning based methods. Since our work belongs to learning based approach, we pay our attention on reviewing learned feature detectors. For more details on classical handcrafted feature detectors, interested readers can refer to a benchmark work from Schmid \etal. \cite{Schmid2004EvaluationOI}.

FAST \cite{Rosten2006MachineLF} is one of the first attemps to apply machine learning technique for reliable and fast corner detection. Similar strategies have also been applied in other related extensions \cite{Rosten2010FasterAB, Rublee2011ORBAE, Leutenegger2011BRISKBR}.
TILDE \cite{Verdie2015TILDEAT}, one of the first deep learning based detector, trains a piece-wise linear regressor to detect keypoints under drastic weather and illuminstion changes, using SIFT keypoints as supervision. 
DetNet \cite{Lenc2016LearningCF} derives a covariant constraint to learn stable anchors for local features. It is further extended by Zhang \etal. \cite{Zhang2017LearningDA} to introduce two new concepts of standard patch and canonical feature for feature detection. 
%
Savinov \etal \cite{Savinov2017QuadNetworksUL} later proposes Quad-networks, which is unsupervised and trains a neural network to rank points under a transformation-invaraint manner. It then extracts keypoints from the top/bottom quantiles of the rankings.
A similar detector is proposed by Zhang \etal \cite{Zhang2018LearningTD}, which combines the same ranking loss with a grid-wise peakness loss to detect keypoints in texture images. 
Key.Net \cite{Laguna2019KeyNetKD} resorts to using hand-crafted filters together with learned convolutional neural network (CNN) features by a light weight CNN network. They propose to use a spatial softmax operator for detecting keypoints across multi-scale regions. 
ELF \cite{Benbihi2019ELFEL} proposes to use a pre-trained CNN for image classification task to detect saliency keypoints without requiring extra training. Recently, it has been shown that convolutional neural network trained for descriptor can also be used for keypoint detection, and achieves impressive performance \cite{Tian2020D2DKE}.
In addition to the above methods which are solely designed for keypoint detection, existing works also seek the possibility to integrate both the feature detection and description in a unified framework. 
For example, LIFT \cite{Yi2016LIFTLI} takes a quadruplet of patches to jointly train a detector, an orientation estimator, and a descriptor which are supervised by the feature correspondences from a Structure-from-Motion (SfM) pipeline. 
%
Instead of getting supervision via a SfM pipeline, SuperPoint \cite{DeTone2018SuperPointSI} proposes to first pre-train the detector via a synthetic dataset, which consists of primitive geometric shapes. They then take advantage of a homographic adaptation module to achieve self-supervised training together with the feature descriptor network on real images. 
LF-Net \cite{Ono2018LFNetLL} proposes to enforce same feature response for corresponding points across images to train the detector. The correspondence is built based on known camera poses and depth maps.
%

Different from existing works, which are usually trained to detect keypoints from a sharp image, we propose a motion blur aware keypoint detector for robust 3D vision under low-lighting conditions.

\PAR{Image deblurring.}
Existing motion deblurring algorithms can be mainly categorized into classical gradient-descent based methods and learning based methods during inference. We will only focus on several representative learning based single image deblurring networks, which are most related to our work.
Early learning based methods \cite{Sun2015LearningAC, Gong2017FromMB, Schuler2016LearningTD, Yan2016BlindIB} mainly exploit deep network to estimate the unknown blur kernels and then employ conventional deconvolution methods to restore the blurred images.  
The performance of single image deblurring algorithms has been further boosted by end-to-end deep neural networks, which formulate the deblurring problem as an image translation problem. 
%
Xu \etal \cite{Xu2014DeepCN} develop a deep convolutional neural netowrk to capture the blur degradation for image deconvolution. 
Sun \etal \cite{Sun2015LearningAC} apply the network to predict the varying motion blur kernels, which enables to image deblurring.
Nah \etal \cite{Nah2017DeepMC} follow a coarse-to-fine approach to train a multi-scale CNN for blind deblurring.
Kupyn \etal propose DeblurGAN \cite{Kupyn2018DeblurGANBM} and DeblurGAN-v2 \cite{Kupyn2019DeblurGANv2D} based on a adversarial loss for motion deblurring. 
%
Tao \etal \cite{Tao2018ScaleRecurrentNF} propose a SRN-DeblurNet, which adopts a scale-recurrent structure to realize multi-scale image deblurring.
Liu \etal \cite{Liu2020SelfSupervisedLM} recently propose a self-supervised network for motion deblurring.
Although those methods achieve impressive performance on image deblurring, they usually require large computational resources and are hard to run in real-time. Instead of firstly deblurring each image and then detecting keypoints on the restored image, we propose an efficient one-stage feature detector from motion blurred image directly.
%

\PAR{MLP-like architecture in computer vision.}
While convolutional neural networks and Vision Transformers (ViT) \cite{Dosovitskiy2021AnII} have been the de-facto standards for many computer vision applications, MLP-based architectures have also achieved state-of-the-art performance in several vision tasks recently \cite{Tolstikhin2021MLPMixerAA, Liu2021PayAT, Yu2021S2MLPv2IS, Tu2022MAXIMMM, Lian2022ASMLPAA, Chen2022CycleMLPAM, Choe2022PointMixerMF, Mansour2022ImagetoImageMF, Tang2022AnIP, Touvron2022ResMLPFN, Guo2022HireMLPVM, Li2022BraininspiredMP, Hou2022VisionPA}. Due to the conceptually and technically simplicity, MLP-based architecture is getting more attention in both visual recognition \cite{Liu2021PayAT, Yu2021S2MLPv2IS} and dense prediction tasks \cite{Lian2022ASMLPAA, Chen2022CycleMLPAM}. Recently, MAXIM \cite{Tu2022MAXIMMM} adopts a multi-axis gated MLP module for low-level image processing while SegFormer \cite{Xie2021SegFormerSA} unifies Transformers with MLP decoders for semantic segmentation tasks. While MLP-based architecture has achieved impressive performance in several vision related tasks, their applicability in local feature detection has never been explored. We thus propose to explore this possibility for local feature detection, and achieve state-of-the-art performance over prior keypoint detectors with both motion blurred and natural sharp images. 

\vspace{-0.5em}

\section{Methods}
\label{sec:methods}
\begin{figure*}
	\centering
	\begin{subfigure}{0.9\linewidth}
		\centering
		\includegraphics[width=0.9\linewidth]{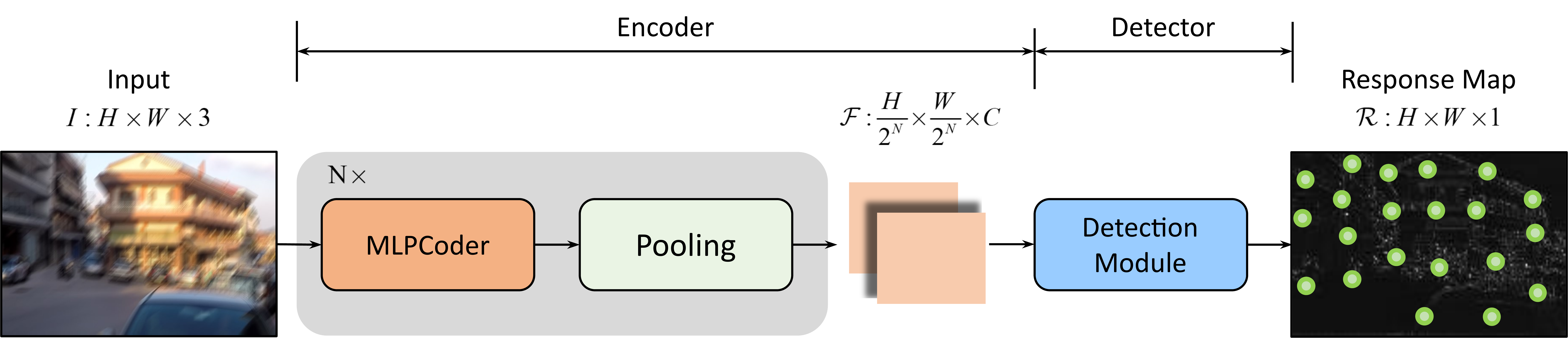}
		\caption{\textbf{BALF Framework}}
		\label{fig_overview}
	\end{subfigure}
	\centering
	\begin{subfigure}{0.325\linewidth}
		\centering
		\includegraphics[width=0.9\linewidth]{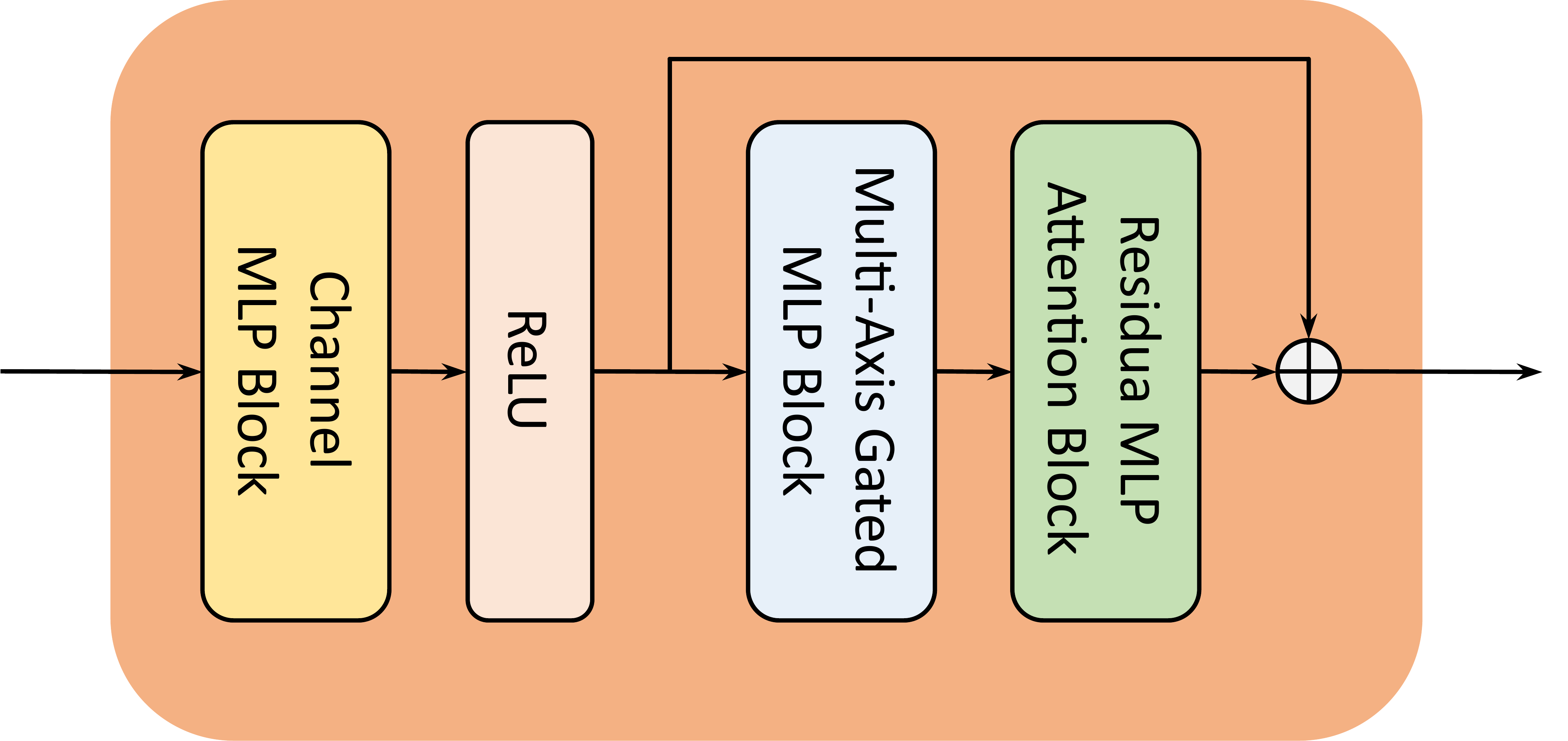}
		\caption{\textbf{MLPCoder Block}}
		\label{fig_mlpcoder}
	\end{subfigure}
	\centering
	\begin{subfigure}{0.325\linewidth}
		\centering
		\includegraphics[width=0.9\linewidth]{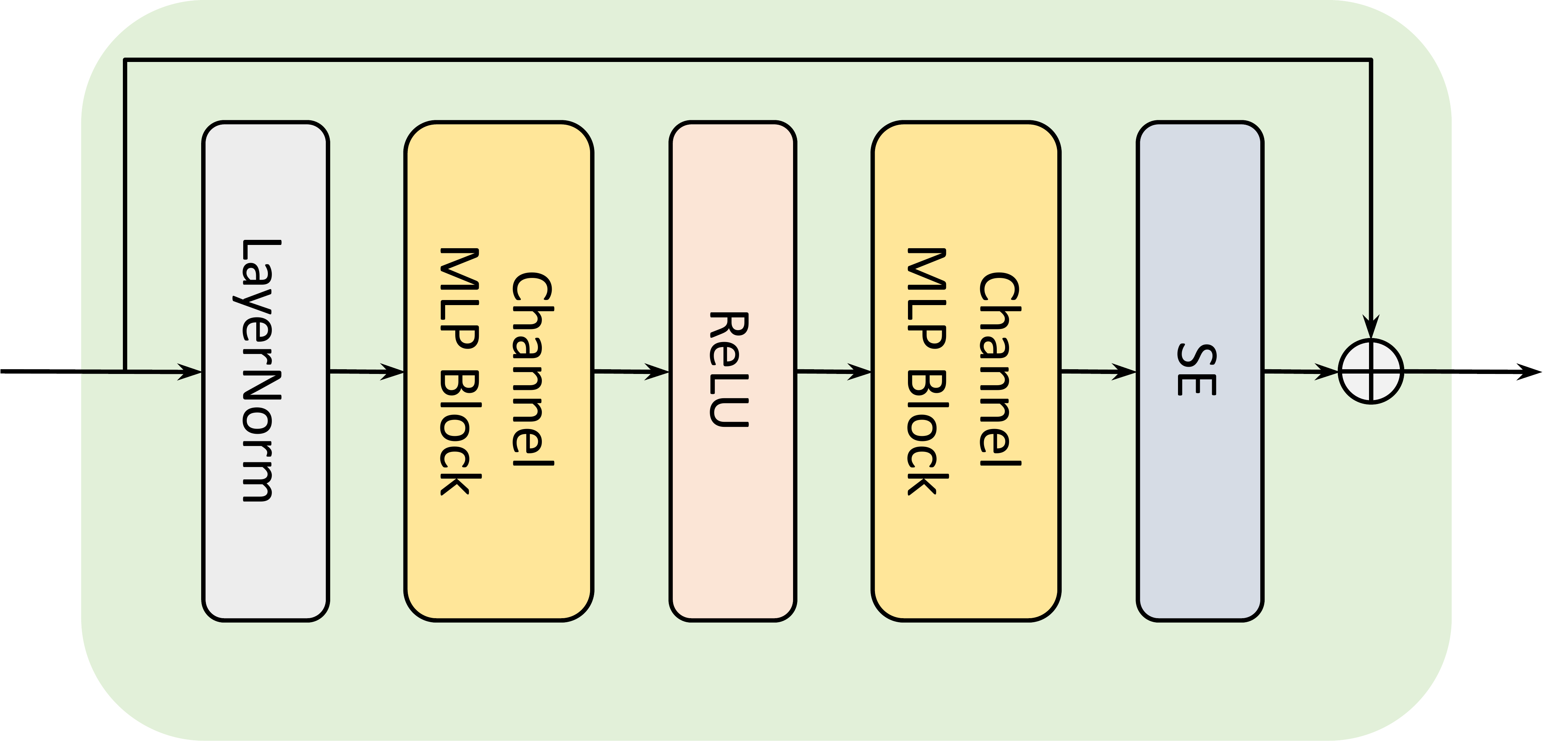}
		\caption{\textbf{Residual MLP Attention Block (RMAB)}}
		\label{fig_rmab}
	\end{subfigure}
	\caption{\textbf{The proposed network for motion blur aware local feature detector (BALF).} Our network consists of two main modules: a multi-stage MLP-based encoder to extract an intermediate feature representation of the input image, and a MLP detection module to detect salient keypoints via a differentiable softmax operator.}
	\label{fig_balf}
	\vspace{-1em}
\end{figure*}

We present, to the best of our knowledge, the first MLP-based architecture (as shown in \cref{fig_balf}) for local feature detection with a blurred image. As shown in \cref{fig_overview}, our network consists of a MLP-based encoder and a MLP-based detection module. The encoder network learns a effective feature representation of the input image via cascaded MLP blocks. The learned feature representation is then input to a detection module, which takes advantages of a differentiable channel-wise softmax operator. The resulting network is light-weight and effective. We will detail each component as follows.

\subsection{MLP-based encoder}
Our MLP-based encoder contains cascaded MLPCoder blocks (as shown in \cref{fig_mlpcoder}) to obtain a pyramid level representation of the input image. Each MLPCoder block contains a channel MLP block, a multi-axis gated MLP block and a residual MLP attention block. The channel MLP block maps each pixel to a high-dimensional representation, such that it can be further processed by the following blocks. After each MLPCoder block, we apply a max-pooling layer to extract the most significant features and reduce the spatial dimension of the feature representation. The loss of the spatial resolution caused by the pooling layer is compensated by the increased number of channels of the feature representation. 

\PAR{Multi-axis gated MLP block.} For keypoint detection task, local region relationship among the pixels is usually more important compared to long-range relationship. However, due to motion blur, the intensity of a particular pixel is mixed by that of several neighboring pixels. It would be beneficial to have a larger receptive field to take account of more context information for better localization of the salient keypoints. We thus take advantages of a state-of-the-art MLP block, \ie the multi-axis gated MLP block, from MAXIM \cite{Tu2022MAXIMMM} to extract both local and global visual cues. MAXIM \cite{Tu2022MAXIMMM} is a MLP-based network for low-level image processing tasks, and achieves state-of-the-art performance on image denoising, deblurring, deraining, dehazing and enhancement compared to prior works. Multi-axis gated MLP block presents a principled way to apply 1D operators on 2D images in a scalable manner, and apply them in parallel corresponds to both local and global (dilated) mixing of spatial information respectively. For more details, interested readers are suggested to refer to the work of Tu \etal \cite{Tu2022MAXIMMM}.

\PAR{Residual MLP attention block.} The multi-axis gated MLP block mainly learns spatial dependencies across the feature representations. To better capture the channel-wise dependencies and inspired by convolutional channel attention block in \cite{Woo2018CBAMCB, Zamir2021MultiStagePI}, we exploit the squeeze-and-excitation (SE) block from SENet \cite{Hu2020SqueezeandExcitationN} to build up a residual MLP attention block (RMAB). SE block is more efficient compared to those convolutional channel attention blocks. It allows the network to perform feature re-calibration and exploit the inter-channel relationship of features, through which it can learn to use global information to selectively emphasize informative features and suppress less useful ones. 

The SE block first conduct a squeeze operation, which produces a channel descriptor by aggregating feature maps across their spatial dimensions. The function of this descriptor is to produce an embedding of the global distribution of channel-wise feature responses, allowing information from the global receptive field of the network to be used by all its layers. The aggregation is followed by an excitation operation, which takes the form of a simple self-gating mechanism that takes the embedding as input and produces a collection of per-channel modulation weights. These weights are applied to the input feature maps to generate the output of the SE block which can be fed directly into subsequent layers of the network. 
We further improve the channel attention concept for keypoint detection task, by applying two channel MLP blocks to build a residual MLP attention block (as shown in \cref{fig_rmab}) together with the SE block \cite{Hu2020SqueezeandExcitationN}.

\begin{figure*}[htb]
\centering
\includegraphics[width=0.9\linewidth]{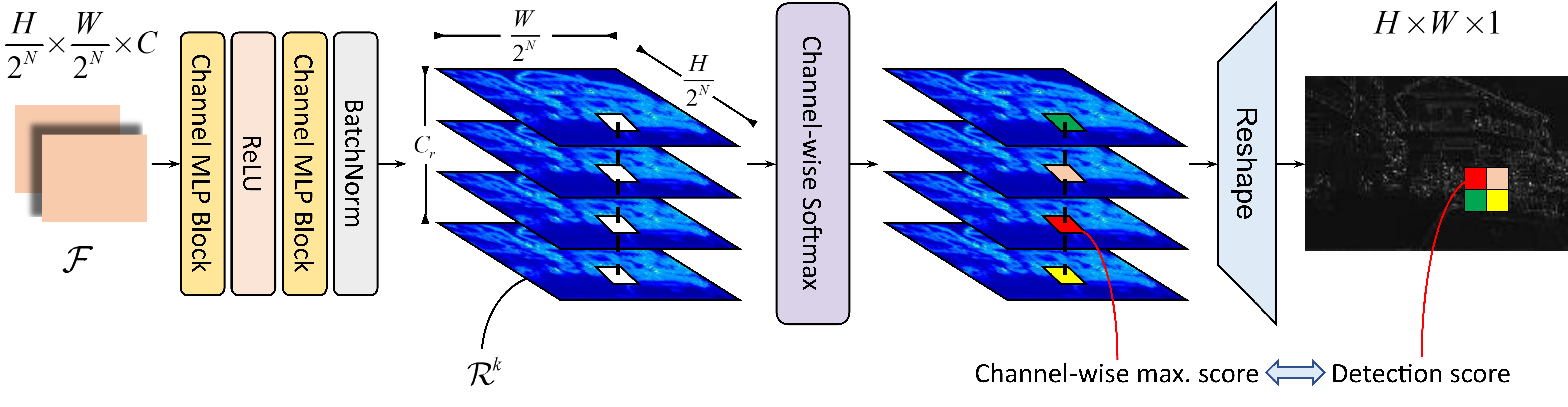}
\caption{\textbf{Detection module.} The learned feature representations of the input image is processed by two channel-wise MLP blocks. The keypoints are then detected by using channel-wise softmax operation and mapped back to the original image domain.}
\label{fig_detection_module}
\vspace{-1em}
\end{figure*}

\subsection{MLP-based detection module}
After the MLP-based encoder, the resulting feature map $\mathcal{F} \in \mathbb{R}^{H^\prime \times W^\prime \times C}$, where $H^\prime = \frac{H}{2^N}$ and $W^\prime = \frac{W}{2^N}$, are passed through a MLP-based detection module (as shown in \cref{fig_detection_module}) to output a dense probability map for each pixel as a keypoint. 

The details of the detection module are illustrated in \cref{fig_detection_module}. It consists of two channel MLP blocks, which transform the input feature representation from $C$ channels to $C_r$ channels at the same spatial resolution, where $C_r = 4^N$. Each element along the channel direction corresponds to the response of a particular pixel of the input full resolution image. In other words, all the elements along the channel direction correspond to the responses of pixels within a $2^N \times 2^N$ sized patch of the input image. We use $N=3$ in our experiments.

Handcrafted feature detectors, such as SIFT \cite{LoweDavid2004DistinctiveIF}, usually perform spatial non-maxima suppression (NMS) on the response map, to select a set of sparsely distributed salient keypoints. To achieve similar goal, we apply a differentiable channel-wise softmax operation on the transformed feature representation as shown in \cref{fig_detection_module}. The corresponding channel index of the maximum responsed element is then mapped to the pixel index of the full resolution input image, such that the corresponding pixel can be assigned with the response score. We further eliminate pixels as keypoints for those with low responses by a pre-defined threshold, such as pixels within a homogeneous region. 
The purpose of using this feature detection procedure is two-fold. First, the channel-wise softmax layer is similar to NMS but is differentiable and enables end-to-end training. Secondly, it replaces the decoder module with simple non-parametric operator. It neither involves any feature learning nor introduces any additional parameters, which further reduces the computational overhead.

\subsection{Loss function}
We formulate the keypoint detection as a regression problem. We train the network with a  ground truth response map $\mathcal{R}_{GT}$, which are generated by detecting SIFT keypoints \cite{LoweDavid2004DistinctiveIF} on the corresponding sharp images and placing Gaussian kernels at those locations. The loss function used to train the network can be formalized described as:
\begin{equation}
    \mathcal{L} = ||\mathcal{R}-\mathcal{R}_{GT}||^2,
\end{equation}
where $\mathcal{R}$ is the predicted response map (as shown in \cref{fig_overview}) and $\mathcal{R}_{GT}$ is the ground truth response map. 

\section{Experiments}
\label{sec:experiments}
The main motivation of our work is to develop a motion blur aware local feature detector via deep networks, since many state-of-the-art methods for a sharp image have been proposed and achieved impressive performance. There is no existing dataset to evaluate local feature detectors on motion blurred images. We thus create a training dataset via a publicly available single image deblurring dataset, \ie the GoPro dataset from Nah \etal \cite{Nah2017DeepMC}, which has paired sharp and blurred images. To evaluate our method, we generate a synthetic dataset via HPatches dataset \cite{Balntas2017HPatchesAB}, which is commonly used for local feature evaluations. HPatches dataset is different from the GoPro dataset, and it can thus also reflect the generalization performance of our method. We therefore mainly use this dataset for evaluation. 

\PAR{Datasets.} We use the GoPro dataset from Nah \etal \cite{Nah2017DeepMC} to generate data to train our network. The GoPro dataset is commonly used for evaluations of single image deblurring networks. It consists of 3,214 paired sharp and blurred images, which are captured from 33 scenes. We follow their convention and use 22 sequences for training and 11 sequences for testing. 
To generate supervision data for each blurred image, we assume the blurred image should have the same keypoint locations as those for its paired sharp image. We use SIFT \cite{LoweDavid2004DistinctiveIF} to first detect keypoints from the sharp image, and then generate a heatmap and place Gaussian kernels as the ground truth response map for network training. 

\begin{figure}
	\setlength\tabcolsep{1pt}
	\begin{tabular}{cccc}
		\includegraphics[width=0.248\columnwidth]{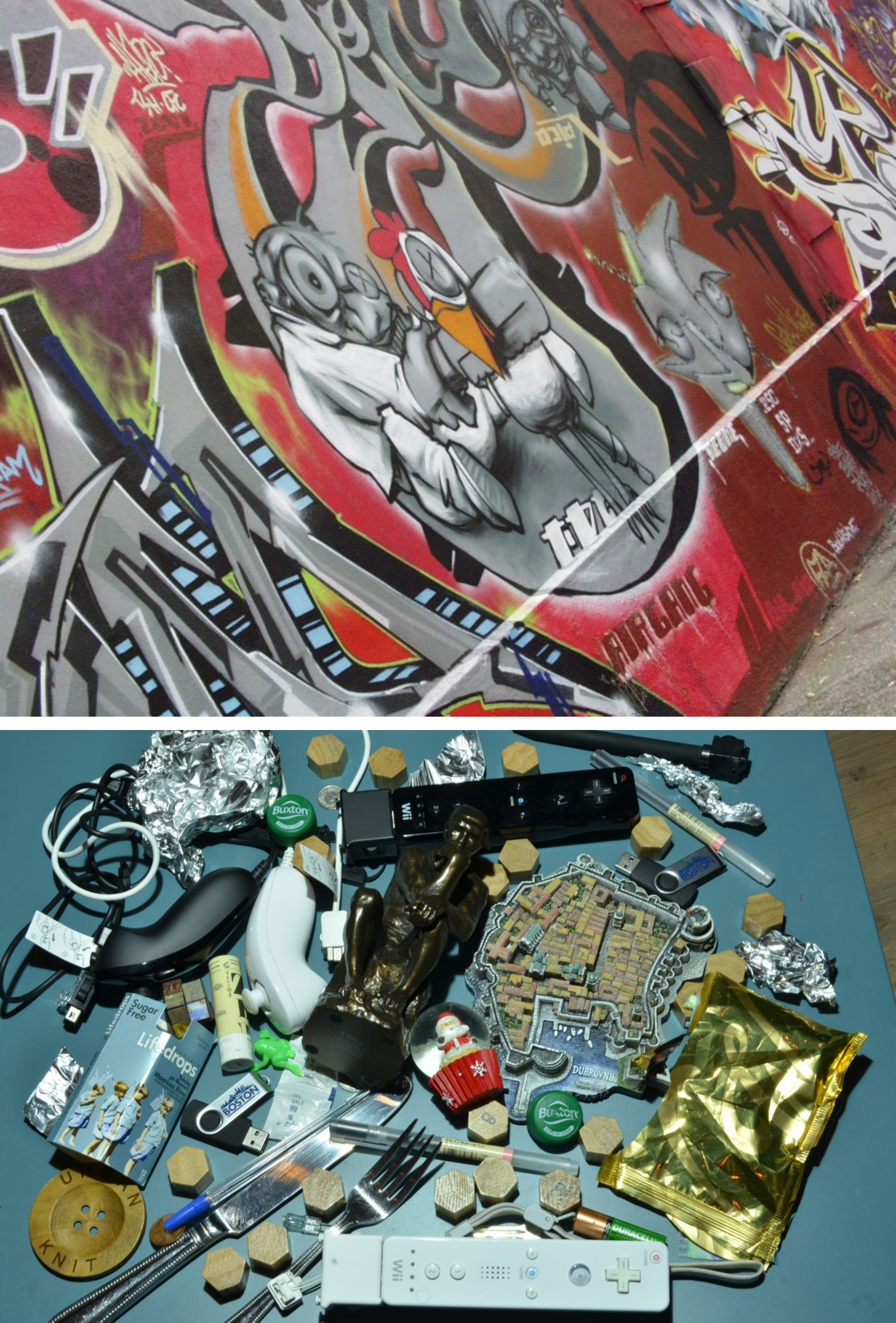} &
		\includegraphics[width=0.248\columnwidth]{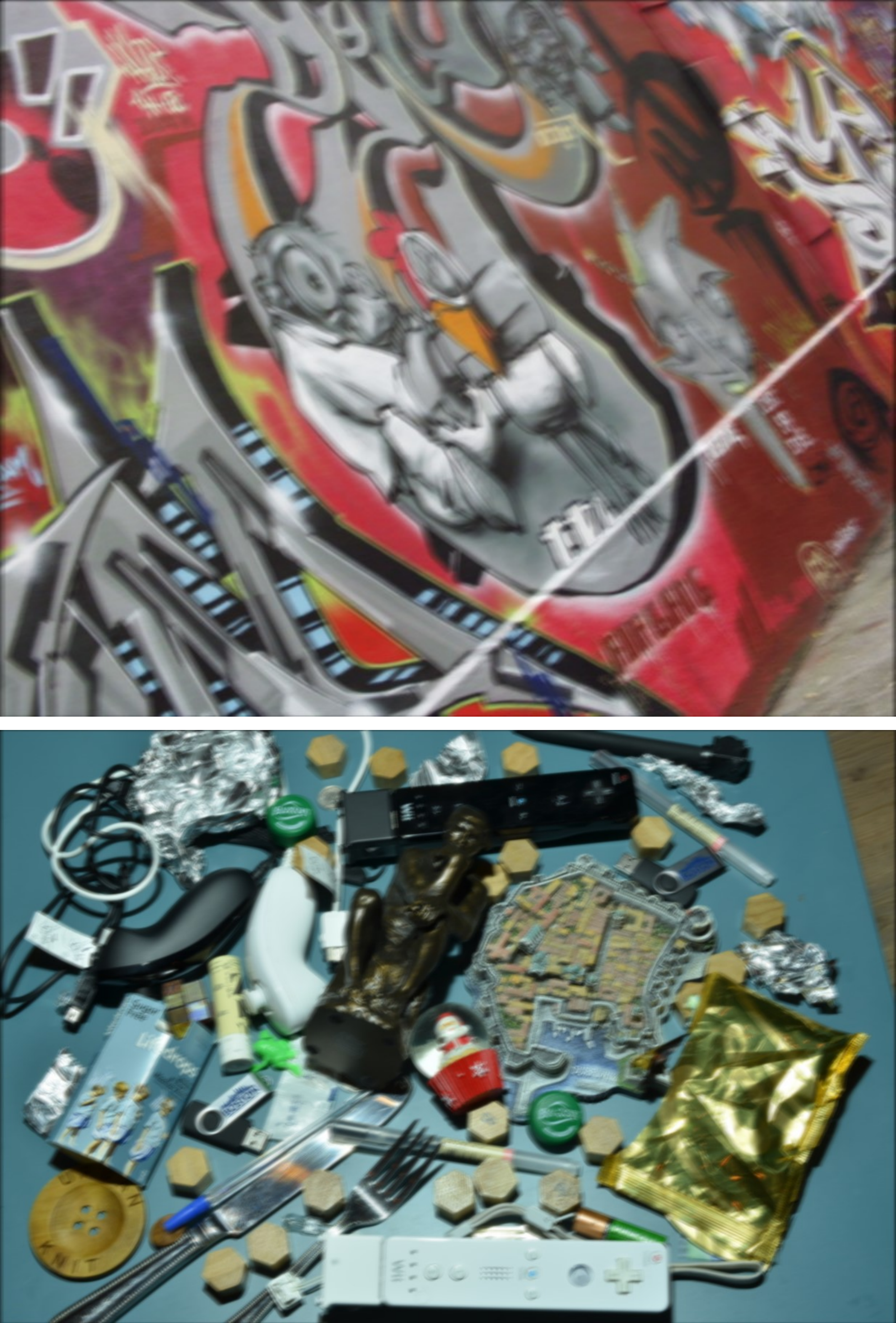} &
		\includegraphics[width=0.248\columnwidth]{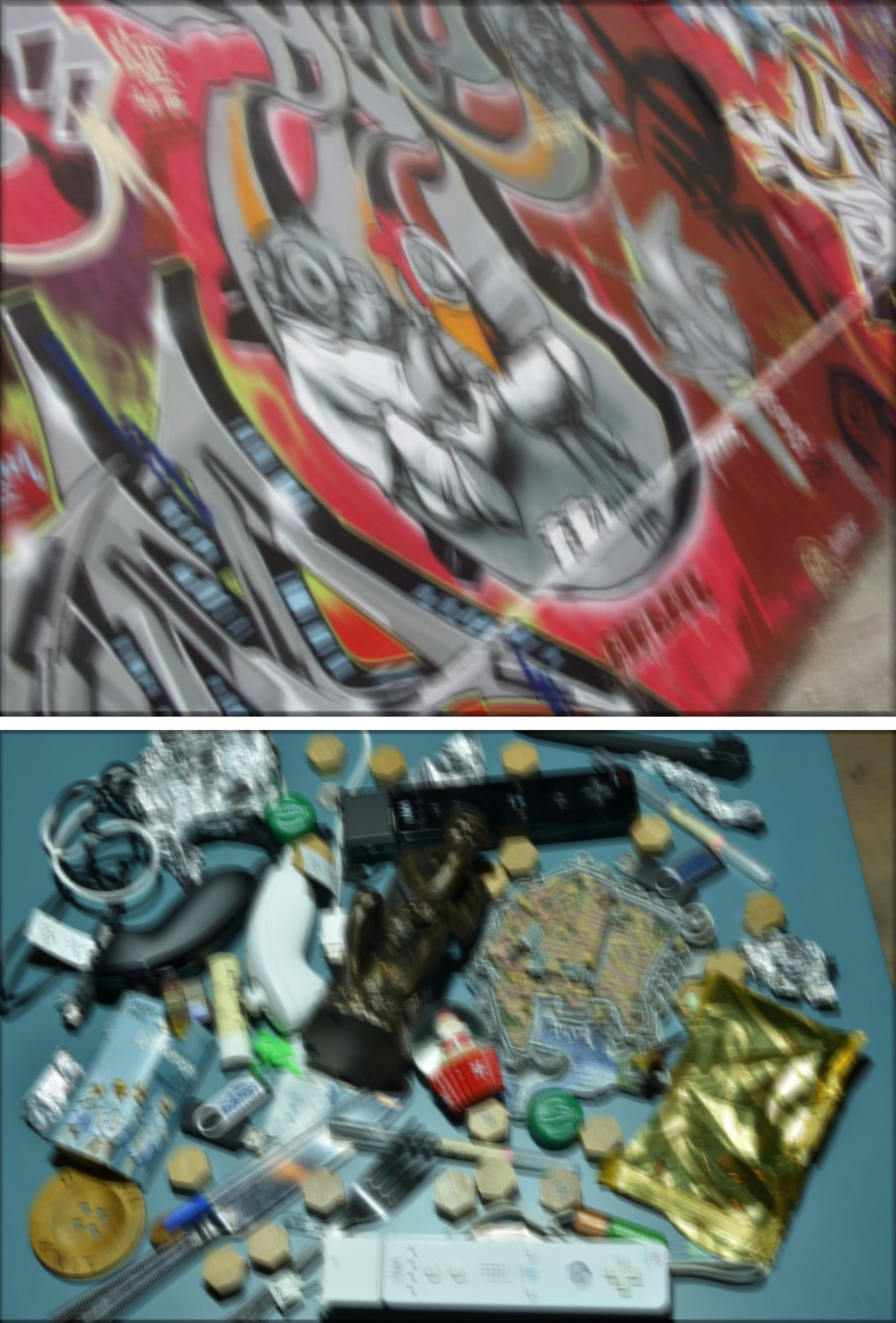} &
		\includegraphics[width=0.248\columnwidth]{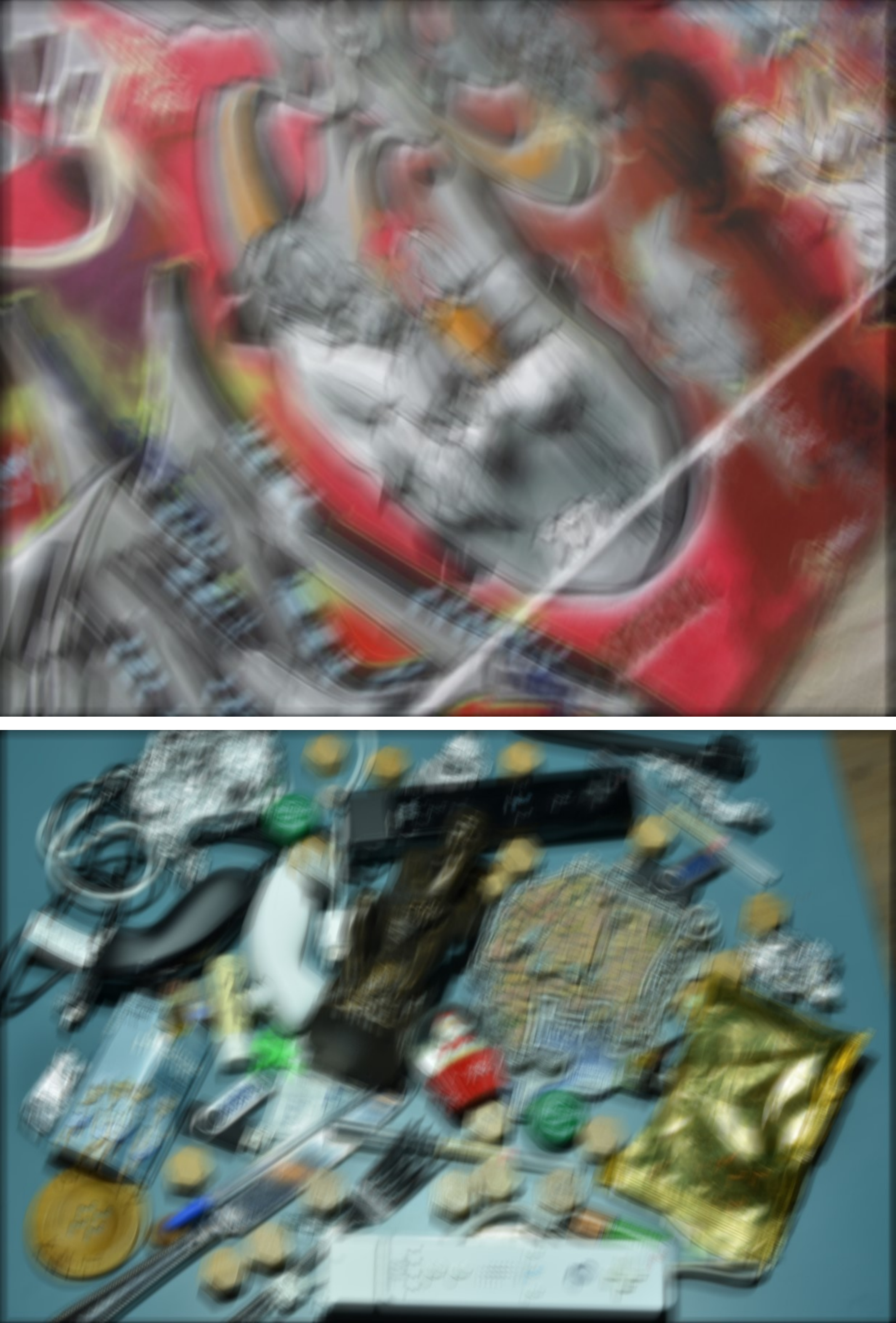} \\
		Sharp & E\begin{footnotesize}ASY\end{footnotesize} & H\begin{footnotesize}ARD\end{footnotesize} & T\begin{footnotesize}OUGH\end{footnotesize}
	\end{tabular}
	\caption{{\textbf{Example synthesized blurred images from HPatches dataset \cite{Balntas2017HPatchesAB}.}} The first column shows the sharp image, and the next three columns are blurred images with varying blur levels. Best viewed in high resolution.}
	\label{fig_blur_hpatches_samples}
	\vspace{-1.2em}
\end{figure}

To better evaluate the generalization performance of our method, we generate a synthetic motion blurred image dataset by using HPatches dataset \cite{Balntas2017HPatchesAB}. HPatches dataset is commonly used for local feature evaluations, which originally covers both illumination and viewpoint changes. It gathers images from existing datasets, such as DTU \cite{Aans2011InterestingIP} and Oxford \cite{Mikolajczyk2005ACO} datasets. It provides a total of 116 sequences and is further divided into 59 sequences for viewpoint changes and 57 sequences for illumination changes. Each sequence includes a reference image and 5 target images with varying viewpoint or illumination changes, together with the corresponding homographies between them. We generate a random motion blur kernel, \ie point spread function, for each image from the HPatches dataset to synthesize a motion blurred image. We generate three varying levels of motion blur, by changing the blur kernel size and motion irregularities (\ie the non-linearity of the motion). \cref{fig_blur_hpatches_samples} presents several example images for different blur levels. We use this dataset purely for evaluation.

\PAR{Implementation details.}
Our network is light-weight and end-to-end trainable. It requires neither large-scale pre-training nor progressive training. During training, we use a batch size of 4, and Adam optimizer with a initial learning rate of $10^{-4}$. After the network is trained for 20 epochs, we linearly decay the learning rate to $10^{-6}$ till the $50^{th}$ epoch. The network takes about 3 hours to be trained on a single NVIDIA Geforce 2080Ti graphic card. During training, we also perform data augmentation as that of \cite{Laguna2019KeyNetKD}. In particular, we perform a sequence of random rotation, random scaling, random skewing and random perspective transformation on the original image. We also apply random photometric transformation, such that the trained network is robust against illumination changes. Then we randomly crop a $256\times 256$ pixels image for training. Detailed network architecture can be found from our supplementary material.  

\PAR{Baseline methods and evaluation metrics.}
We compare our approach against a number of representative detectors, which range from classical handcrafted methods to recent learning based methods. In particular, we compare against SIFT \cite{LoweDavid2004DistinctiveIF}, SURF \cite{Bay2006SURFSU}, Harris Laplace \cite{Mikolajczyk2004ScaleA}, Shi-Tomasi \cite{Shi1994GoodFT}, MSER \cite{Matas2004RobustWS}, KAZE \cite{Alcantarilla2012KAZEF}, AKAZE \cite{Alcantarilla2013FastED} and FAST \cite{Rosten2010FasterAB} with their OpenCV implementations. We also compare with learned feature detectors, such as LIFT \cite{Yi2016LIFTLI}, Key.Net \cite{Laguna2019KeyNetKD}, SuperPoint \cite{DeTone2018SuperPointSI}, LF-Net \cite{Ono2018LFNetLL}, D2-Net \cite{Dusmanu2019D2NetAT}, and R2D2 \cite{Revaud2019R2D2RA}. For fair comparisons with motion blurred images, we also re-train those learning based methods on the GoPro dataset using their publicly available source codes. For evaluations with sharp images, we still use their official provided pretrained models. 

The repeatability metric proposed in \cite{Mikolajczyk2003APE} measures the quality of keypoint detection and is commonly used by the community. For a pair of images, it is computed as the ratio between the number of corresponding keypoints observed by both images and the smaller number of keypoints detected in one of the two images. To identify the corresponding keypoints, we compute the overlap error, $\epsilon_{IoU}$, between the regions of two candidate keypoints as in \cite{Zhang2017LearningDA, Zhang2018LearningTD}. We also fix the maximum number of extracted keypoints and allow each keypoint to be matched only once as in \cite{Yi2016LIFTLI} for fair evaluations. In our experiments, we consider the top 1000 interest points for repeatability computation. It is considered as a correct match if $\epsilon_{IoU}$ is smaller than 0.4, \ie, the overlap between corresponding region is more than $60\%$.

\PAR{Ablation study}
In this section, we study the effect of the residual MLP attention block (RMAB). We conduct experiments with three settings, \ie the network with and without RMAB, and we also replace RMAB with a residual convolutional attention block (RCAB) from \cite{Woo2018CBAMCB}. The experimental results shown in \cref{table_rmab_effects} demonstrate that RMAB block is indeed effective for local feature detection. The RCAB block requires more parameters, longer inference time and achieves lower repeatability compared to the network with RMAB. It further demonstrates the potential to use MLP blocks for efficient network architectures. The experiments are conducted with the GoPro testing sequences.

\begin{table}
\centering
\setlength{\tabcolsep}{5pt}
\resizebox{0.99\linewidth}{!}{
    \begin{tabular}{lccc}
        \toprule
        Variant & Repeatability $\uparrow$ & Params  $\downarrow$ & Inference time  $\downarrow$ \\
        \midrule
        w/o RMAB & 63.59\% & 326K & 19.18ms \\
        RCAB & 73.56\% & 725K & 57.29ms\\
        RMAB (proposed) & 75.15\% & 381K & 29.02ms\\
        \bottomrule
        \end{tabular}
    }
    \caption{\textbf{Ablation study of the residual MLP attention block (RMAB).} The experimental results demonstrate that the RMAB block is indeed efficient and effective for local feature detection.}
    \label{table_rmab_effects}
    \vspace{-1.2em}
\end{table}


\begin{table*}
    \begin{minipage}{.38\linewidth}
      \centering
        \setlength{\tabcolsep}{3.6pt}
        \resizebox{1.0\linewidth}{!}{
    	\begin{tabular}{lcc|c}
    		\toprule
    		& \multicolumn{3}{c}{\begin{tabular}[c]{@{}c@{}}Reference: Sharp\\Target: Sharp\end{tabular}}\\
    		\cmidrule{2-4}
    		Method & Viewpoint  $\uparrow$ & Illumination  $\uparrow$ & Total  $\uparrow$
    		\\ \midrule
    		SIFT \cite{LoweDavid2004DistinctiveIF} & 60.29 & 60.44 & 60.36\\
    		SURF \cite{Bay2006SURFSU} & 62.67 & 64.01 & 63.33\\
    		Harris-Laplace \cite{Mikolajczyk2004ScaleA} & 63.89 & 62.91 & 63.41\\
    		Shi-Tomasi \cite{Shi1994GoodFT} & \underline{69.28} & 64.13 & 66.74\\
    		MSER \cite{Matas2004RobustWS} & 52.45 & 50.58 & 51.53\\
    		KAZE \cite{Alcantarilla2012KAZEF} & 67.30 & 65.67 & 66.50\\
    		AKAZE \cite{Alcantarilla2013FastED} & 66.08 & 69.07 & 67.55\\
    		FAST \cite{Rosten2010FasterAB} & 66.08 & 63.65 & 64.88\\
    		LIFT \cite{Yi2016LIFTLI} & 56.97 & 60.73 & 58.82 \\
    		Key.Net \cite{Laguna2019KeyNetKD} & 68.99 & 67.47 & 68.24\\
    		SuperPoint \cite{DeTone2018SuperPointSI} & \textbf{69.53} & 68.92 & 69.23\\
    		LF-Net \cite{Ono2018LFNetLL} & 68.41 & \textbf{73.61} & \textbf{70.96}\\
    		D2-Net \cite{Dusmanu2019D2NetAT} & 53.99 & 62.80 & 58.32\\
    		R2D2 \cite{Revaud2019R2D2RA} & 61.68 & 61.93 & 61.80 \\
    		\midrule
    		\begin{tabular}[c]{@{}l@{}}BALF (ours)\end{tabular} & 67.21 & \underline{73.51} & \underline{70.28}\\
    		\bottomrule
    		\end{tabular}
    	}
	\caption{\textbf{Repeatability results (\%) on HPatches \cite{Balntas2017HPatchesAB} sharp image pairs.} For each method, we report the average repeatability score on the viewpoint change, illumination change, and all image sequences.}
	\label{table_sharp_hpatches}
    \end{minipage}
    \hfill
    \begin{minipage}{.59\linewidth}
      \centering
        \setlength{\tabcolsep}{6.388pt}
        \resizebox{1.0\linewidth}{!}{
        	\begin{tabular}{lccclccc}
        	\toprule 
        	& \multicolumn{4}{c}{\begin{tabular}[c]{@{}c@{}}Reference: Sharp\\Target: Blur\end{tabular}}
        	& \multicolumn{3}{c}{\begin{tabular}[c]{@{}c@{}}Reference: Blur\\Target: Blur\end{tabular}}\\
        	\cmidrule{2-4} \cmidrule{6-8}
        	Method & E\begin{footnotesize}ASY\end{footnotesize} $\uparrow$ & H\begin{footnotesize}ARD\end{footnotesize} $\uparrow$ & T\begin{footnotesize}OUGH\end{footnotesize} $\uparrow$ && E\begin{footnotesize}ASY\end{footnotesize} $\uparrow$ & H\begin{footnotesize}ARD\end{footnotesize} $\uparrow$ & T\begin{footnotesize}OUGH\end{footnotesize} $\uparrow$ \\
        	\midrule
        	SIFT\cite{LoweDavid2004DistinctiveIF}& 55.92 & 56.80 & 53.49 &&56.99 & 53.49 & 45.94\\
        	SURF\cite{Bay2006SURFSU}& 58.88 & 56.23 & 56.24 &&61.08 & 58.04 & 53.60\\
        	Harris-Laplace\cite{Mikolajczyk2004ScaleA}& 36.70 & 37.97 & 34.98 &&35.76 & 31.95 & 27.47\\
        	Shi-Tomasi\cite{Shi1994GoodFT}& 57.33 & 55.11 & 49.11 &&56.29 & 53.75 & 51.37\\
        	MSER\cite{Matas2004RobustWS}& 44.19 & 41.97 & 37.05 && 41.81 & 38.24 & 34.59\\
        	KAZE\cite{Alcantarilla2012KAZEF} & 49.90 & 46.84 & 39.98 && 63.29 & 58.71 & 46.90\\
        	AKAZE \cite{Alcantarilla2013FastED}& 54.15 & 50.51 & 45.49 &&\underline{65.16} & \underline{62.20} & 51.54\\
        	FAST\cite{Rosten2010FasterAB} & 61.98 & 61.77 & 51.37 && 57.84 & 53.35 & 51.17\\
        	LIFT\cite{Yi2016LIFTLI} & 50.69 & 50.17 & 46.99 &&48.34 & 46.57 & 46.53\\
        	Key.Net \cite{Laguna2019KeyNetKD}& 60.34 & 54.71 & 44.69 &&62.77 & 58.17 & 49.25\\
        	SuperPoint \cite{DeTone2018SuperPointSI}& \underline{65.64} & \underline{62.22} & 52.84 &&58.60 & 50.03 & 43.28 \\
        	LF-Net \cite{Ono2018LFNetLL} & 63.54 & 61.19 & \underline{56.78} &&60.45 & 59.07 & \underline{57.71}\\
        	D2-Net \cite{Dusmanu2019D2NetAT} & 49.71 & 47.30 & 44.32 &&51.80 & 51.05 & 50.53\\
        	R2D2 \cite{Revaud2019R2D2RA}& 57.99 & 51.73 & 40.57 && 57.49 & 55.31 & 46.86 \\
        	\midrule
        	BALF (ours)& \textbf{74.12} & \textbf{74.45} & \textbf{71.84} &&\textbf{70.48} & \textbf{68.43} & \textbf{67.71}\\
        	\bottomrule
        	\end{tabular}
	}
	\caption{\textbf{Repeatability results (\%) on Blur-HPatches datasets.} Our method achieves the best performance on both blur-to-sharp and blur-to-blur configurations. For compactness, we only report the total repeatability. Detailed results on the
    respective changes can be found in our supplementary material.}
	\label{table_blur_HPatches}
    \end{minipage}
    \vspace{-1.7em}
\end{table*}

\PAR{Evaluation with the original HPatches dataset.} 
To study the performance of our method on sharp images, we compare it against other methods on the original HPatches dataset \cite{Balntas2017HPatchesAB}. For this experiment, our network is trained on the GoPro dataset with both sharp and blurred images, while we use the official pre-trained netwroks for other methods. Those pre-trained models are usually well trained with a much larger dataset, such as the COCO dataset \cite{Lin2014MicrosoftCC} or the ImageNet dataset \cite{ILSVRC15}. The experimental results are presented in \cref{table_sharp_hpatches}. It can be demonstrated that our method performs on-par with the state-of-the-art method (\ie LF-Net \cite{Ono2018LFNetLL}) for local feature detection with sharp images. Our method is only slightly worse (\ie $\sim$0.7\% drop with the repeatability metric) compared to LF-Net \cite{Ono2018LFNetLL}, although our network is trained with a relatively small dataset which contains both sharp and blurred images. The experimental results further demonstrate that our detector (\ie BALF) has superior performance even it is designed and trained to be robust against motion blur.

\begin{table*}[h]
    \vspace{1.5em}
	\centering
	\setlength{\tabcolsep}{5pt}
	\resizebox{0.99\linewidth}{!}{%
		\begin{tabular}{lccclccclccclccc}
			\toprule
			& \multicolumn{7}{c}{\begin{tabular}[c]{@{}c@{}}Reference: Sharp\\Target: Deblur\end{tabular}} 
			&& \multicolumn{7}{c}{\begin{tabular}[c]{@{}c@{}}Reference: Deblur\\Target: Deblur\end{tabular}} 
			\\
			\cmidrule{2-8} \cmidrule{10-16}
			& \multicolumn{3}{c}{SRN-DeblurNet \cite{Tao2018ScaleRecurrentNF}} 
			&& \multicolumn{3}{c}{DeblurGAN-v2 \cite{Kupyn2019DeblurGANv2D}} 
			&& \multicolumn{3}{c}{SRN-DeblurNet \cite{Tao2018ScaleRecurrentNF}}
			&& \multicolumn{3}{c}{DeblurGAN-v2 \cite{Kupyn2019DeblurGANv2D}}
			\\ \cmidrule{2-4} \cmidrule{6-8} \cmidrule{10-12} \cmidrule{14-16}
			Method & 
			E\begin{footnotesize}ASY\end{footnotesize} $\uparrow$ & H\begin{footnotesize}ARD\end{footnotesize}  $\uparrow$& T\begin{footnotesize}OUGH\end{footnotesize} $\uparrow$ &&
			E\begin{footnotesize}ASY\end{footnotesize}  $\uparrow$& H\begin{footnotesize}ARD\end{footnotesize} $\uparrow$ & T\begin{footnotesize}OUGH\end{footnotesize} $\uparrow$ &&
			E\begin{footnotesize}ASY\end{footnotesize} $\uparrow$ & H\begin{footnotesize}ARD\end{footnotesize} $\uparrow$ & T\begin{footnotesize}OUGH\end{footnotesize} $\uparrow$ &&
			E\begin{footnotesize}ASY\end{footnotesize}  $\uparrow$& H\begin{footnotesize}ARD\end{footnotesize} $\uparrow$ & T\begin{footnotesize}OUGH\end{footnotesize} $\uparrow$
			\\
			\midrule
			SIFT \cite{LoweDavid2004DistinctiveIF} &
			56.62 & 55.36 & 53.83 &&
			57.63 & 56.52 & 56.50 &&
			59.75 & 58.13 & 50.63 &&
			59.44 & 57.98 & 51.21
			\\
			SURF \cite{Bay2006SURFSU} &
			61.89 & 59.13 & 54.88 &&
			61.97 & 59.57 & 56.34 &&
			62.44 & 61.26 & 55.27 &&
			62.07 & 60.81 & 55.09
			\\
			Harris-Laplace \cite{Mikolajczyk2004ScaleA}  &
			17.15 & 16.87 & 20.54 &&
			16.60 & 16.90 & 20.24 &&
			36.98 & 35.73 & 32.23 &&
			37.09 & 35.97 & 31.54
			\\
			Shi-Tomasi \cite{Shi1994GoodFT} & 
			60.56 & 56.87 & 48.78 &&
			61.75 & 59.10 & 51.56 &&
			63.18 & 61.03 & 50.88 &&
			63.58 & 61.89 & 53.76
			\\
			MSER \cite{Matas2004RobustWS} & 
			46.65 & 43.23 & 37.90 &&
			47.62 & 45.14 & 40.70 &&
			47.70 & 45.40 & 37.49 &&
			47.84 & 45.56 & 38.01
			\\
			KAZE \cite{Alcantarilla2012KAZEF} & 
			65.14 & 63.10 & 60.16 &&
			65.23 & 63.18 & 61.41 &&
			64.20 & 62.45 & 53.41 &&
			64.13 & 61.87 & 54.19
			\\
			AKAZE \cite{Alcantarilla2013FastED} & 
			66.03 & 64.02 & 60.64 &&
			66.29 & 64.50 & \underline{62.72} &&
			65.71 & \underline{64.08} & 56.10 &&
			65.75 & \underline{63.75} & 57.35
			\\
			FAST \cite{Rosten2010FasterAB} & 
			61.77 & 59.67 & \underline{61.60} &&
			62.00 & 60.44 & 58.74 &&
			62.72 & 61.14 & 50.61 &&
			63.40 & 61.70 & 55.43
			\\
			LIFT \cite{Yi2016LIFTLI} & 
			54.98 & 52.64 & 46.75 &&
			56.59 & 53.54 & 49.09 &&
			55.88 & 53.64 & 45.31 &&
			56.68 & 55.31 & 50.44
			\\
			Key.Net \cite{Laguna2019KeyNetKD} &
			63.28 & 58.01 & 47.10 &&
			63.99 & 59.16 & 49.35 &&
			62.86 & 60.44 & 50.74 &&
			62.73 & 60.58 & 52.96
			\\
			SuperPoint \cite{DeTone2018SuperPointSI} &
			\underline{67.72} & \underline{64.05} & 55.26 &&
			\underline{67.95} & \underline{65.86} & 58.22 &&
			\underline{66.38} & 63.16 & 49.52 &&
			\underline{66.50} & 63.71 & 52.09
			\\
			LF-Net \cite{Ono2018LFNetLL} & 
			62.22 & 59.90 & 54.73 &&
			62.59 & 60.24 & 54.81 &&
			63.06 & 62.03 & \underline{57.28} &&
			63.00 & 61.79 & \underline{57.85}
			\\
			D2-Net \cite{Dusmanu2019D2NetAT} &
			51.81 & 49.49 & 45.94 &&
			52.64 & 50.21 & 45.88 &&
			53.60 & 53.00 & 50.93 &&
			53.93 & 53.29 & 50.74
			\\
			R2D2 \cite{Revaud2019R2D2RA} &
			60.31 & 55.43 & 43.26 &&
			60.46 & 55.68 & 45.38 &&
			58.11 & 54.80 & 45.77 &&
			57.95 & 55.03 & 47.86
			\\
			\midrule \midrule
			BALF (ours) &
			\multicolumn{7}{c}{\begin{tabular}[c]{@{}c@{}}\textbf{74.12} / \textbf{74.45} / \textbf{71.84}\\(E\begin{footnotesize}ASY\end{footnotesize} / H\begin{footnotesize}ARD\end{footnotesize} / T\begin{footnotesize}OUGH\end{footnotesize})\end{tabular}} &&
			\multicolumn{7}{c}{\begin{tabular}[c]{@{}c@{}}\textbf{70.48} / \textbf{68.43} / \textbf{67.71}\\(E\begin{footnotesize}ASY\end{footnotesize} / H\begin{footnotesize}ARD\end{footnotesize} / T\begin{footnotesize}OUGH\end{footnotesize})\end{tabular}}
			\\
			\bottomrule
		\end{tabular}%
	}
	\caption{\textbf{Repeatability results (\%) on deblurred images.} The experimental results demonstrate that single image deblurring network can indeed help with the local feature detection. However, it still cannot perform on-par with our one-stage detection network without doing any intermediate deblurring operation.}
	\label{table_deblur_blur_hpatches}
	\vspace{-1em}
\end{table*}

\PAR{Evaluation with the Blur-HPatches dataset.} To evaluate the performance of our network with motion blurred images, we evaluate it against other methods on the synthesized blurred HPatches dataset. For fair comparisons, we re-train all learning based methods on the same GoPro dataset. For compactness, we report the total repeatability instead of the separated results for both viewpoint and illumination changes as in \cref{table_blur_HPatches}. Detailed results on the respective changes can be found in our supplementary material. We also simulate two different application scenarios, \ie the evaluations with blur-to-sharp and blur-to-blur configurations. The blur-to-sharp configuration can be applied for visual localization task. For example, we can pre-built a large-scale 3D map with high-quality sharp images. It might happen that we would use blurred image to query its location against the 3D map if we walk around with an AR device at night. The blur-to-blur configuration can be applied for visual odometry task, from which all the captured images within a time window are motion blurred. 

The experimental results shown in \cref{table_blur_HPatches} demonstrate that prior methods have degraded detection performance when the images are motion blurred. The performance degrades more as the motion blur becomes severer. However, our method achieves impressive performance compared to prior works. The reason might be that prior learning based methods are usually built based on simple convolutional layers (\eg SuperPoint \cite{DeTone2018SuperPointSI}) or networks which are not specially designed for image deblurring. Motion blurred images thus challenge those networks on keypoint detection task. In contrary, our network is built based on the multi-axis gated MLP block, which has been demonstrated to be effective for low-level image processing, \eg image deblurring \cite{Tu2022MAXIMMM}.

\PAR{Evaluation with the GoPro dataset.}
%
We also evaluate the performance of our network against the other methods on the GoPro testing sequences. For fair comparisons, all learning based methods are re-trained on the GoPro dataset. Since there is no ground truth homographies for GoPro dataset as those of HPatches dataset \cite{Balntas2017HPatchesAB}, we randomly warp each testing image to get paired transformed image for repeatability evaluation. The experimental results presented in \cref{table_blur_GoPro} demonstrate that our network achieves superior performance compared to other approaches. 

\PAR{Evaluation with the Blur-HPatches dataset preprocessed by deblurring network.}
We also study if the deblurring networks can help with keypoint detection from blurred image even they usually cannot run in real-time. In particular, we apply two state-of-the-art deblurring networks, \ie SRN-DeblurNet \cite{Tao2018ScaleRecurrentNF} and DeblurGAN-v2 \cite{Kupyn2019DeblurGANv2D}, to deblur the images from the Blur-HPatches dataset and then evaluate all the other methods on the restored images. We use the official pretrained models and apply them to images from the Blur-HPatches dataset preprocessed by these two deblurring networks without any finetuning.

The experimental results shown in \cref{table_deblur_blur_hpatches} demonstrate that the deblurring networks could indeed help a bit for keypoint detection from blurred image. For example, it improves SuperPoint \cite{DeTone2018SuperPointSI} from 62.22\% to 64.05\% on the repeatability metric for the hard blur-to-sharp case. However, it still cannot outperform our network, which has 74.45\% repeatability on the blurred image directly. The reason might be that single image deblurring networks have limitations to restore image from severe motion blurred image. It further demonstrates that to design an one-stage keypoint detector from blurred image directly, would be a better option compared to that of detecting keypoints from the intermediate deblurred image.

\begin{table}
	\centering
	\setlength{\tabcolsep}{10pt}
	\resizebox{0.99\linewidth}{!}{
		\begin{tabular}{lcc}
			\toprule
			Method & \begin{tabular}[c]{@{}c@{}}Reference: Sharp\\ Target: Blur\end{tabular}  $\uparrow$
			&  \begin{tabular}[c]{@{}c@{}}Reference: Blur\\ Target: Blur\end{tabular} $\uparrow$ \\
			\midrule
			SIFT \cite{LoweDavid2004DistinctiveIF} & 60.53 & 60.03\\
			SURF \cite{Bay2006SURFSU} & 56.49 & 60.03\\
			Harris-Laplace \cite{Mikolajczyk2004ScaleA} & 23.54 & 16.35\\
			Shi-Tomasi \cite{Shi1994GoodFT} & 51.26 & 57.61\\
			MSER \cite{Matas2004RobustWS} & 46.54 & 43.09\\
			KAZE \cite{Alcantarilla2012KAZEF} & 49.35 & 48.86\\
			AKAZE \cite{Alcantarilla2013FastED} & 56.34 & 50.51\\
			FAST \cite{Rosten2010FasterAB} & 51.20 & 45.04\\
			LIFT \cite{Yi2016LIFTLI} & 48.61 & 50.56\\
			LF-Net \cite{Ono2018LFNetLL}  & \underline{60.82} & \underline{66.60}\\
			Key.Net \cite{Laguna2019KeyNetKD} & 57.54 & 58.37\\
			SuperPoint \cite{DeTone2018SuperPointSI} & 53.83 & 51.38\\
			D2-Net \cite{Dusmanu2019D2NetAT} & 53.37 & 56.96\\
			R2D2 \cite{Revaud2019R2D2RA} & 50.34 & 53.94 \\
			\midrule
			BALF (ours) & \textbf{75.68} & \textbf{75.15} \\
			\bottomrule
		\end{tabular}
	}
	\caption{\textbf{Repeatability results (\%) on GoPro testing dataset.} The experimental results demonstrate that our network also achieves the state-of-the-art performance compared to prior works on the GoPro dataset.}
	\label{table_blur_GoPro}
	\vspace{-1.2em}
\end{table}

\PAR{Efficiency and performance with real blurred images.}
We also evaluate the efficiency of our method against other methods. \cref{table_efficiency} presents the computational time for all those keypoint detectors. The handcrafted detectors are evaluated on a CPU (Intel i7-8700), and remaining learning based methods\footnote{Since it is impossible to separate the detector and descriptor networks for individual evaluations from D2-Net or R2D2, we did not measure their time consumption.} run on a NVIDIA Geforce 2080 Ti. For fair comparisons, we remove the descriptor network for evaluation from LIFT, SuperPoint, and LF-Net. The experimental results demonstrate that our motion blur aware detector is able to run in real-time ($\sim$34.46 FPS) with a VGA resolution image (\ie 480$\times$640 pixels). It further demonstrates that our detector can be applied to many time-constrained applications, such as robotic visual navigation. 

\begin{figure}[!ht]
	\centering
	\includegraphics[width=0.99\columnwidth]{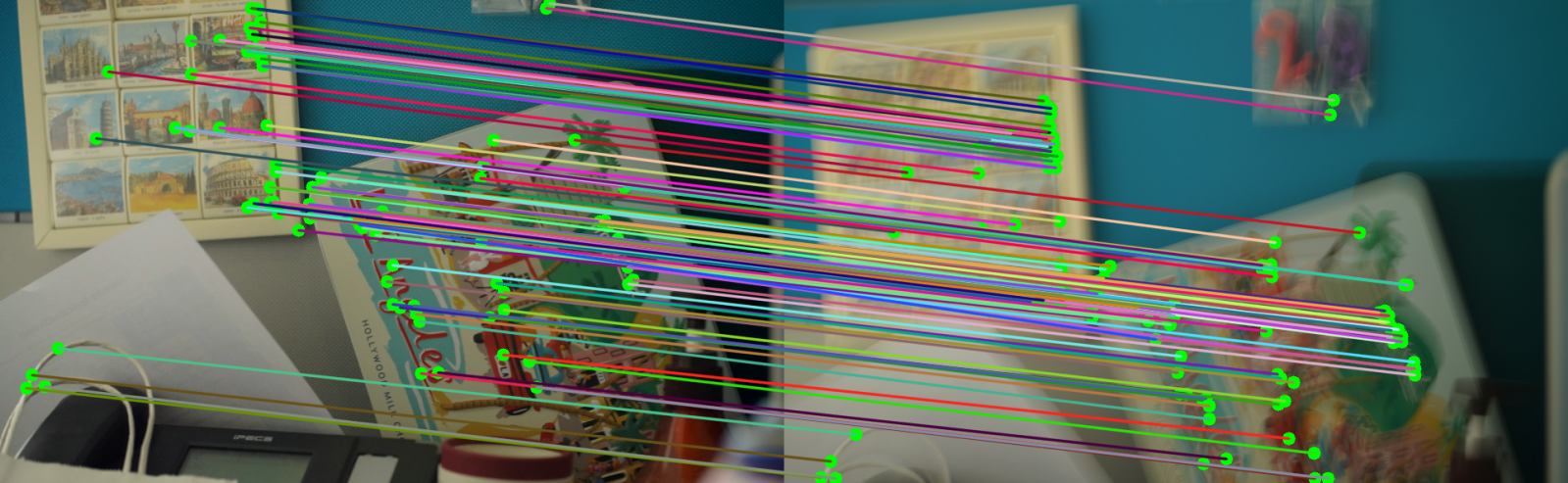}
	\caption{\textbf{Qualitative evaluations on real blurred image.} The experimental results demonstrate that our network is able to detect well distributed and localized keypoints from either sharp and blurred images for further image matching.}
	\label{fig_real}
	\vspace{-0.5em}
\end{figure}

To further demonstrate the performance of our network on the real blurred images, we also present a qualitative feature matching result between a sharp image and a blurred image in \cref{fig_real}. The images are selected from the RealBlur dataset \cite{Rim2020RealWorldBD}, which are captured by real cameras. It demonstrates that our network is able to detect well localized and repeatable keypoints from both sharp and blurred images. More details on the qualitative results can be found from our supplementary material.

\begin{table}
	\footnotesize
	\setlength{\tabcolsep}{8pt}
	\centering
	\resizebox{0.99\linewidth}{!}{
		\begin{tabular}{lcc}
			\toprule
			Method & 240$\times$320 pixels  $\downarrow$ & 480$\times$640 pixels  $\downarrow$\\
			\midrule
			SIFT \cite{LoweDavid2004DistinctiveIF}        & 21.80 & 66.70 \\
			SURF \cite{Bay2006SURFSU}        & 148.46 & 165.78 \\
			Harris-LapLace \cite{Mikolajczyk2004ScaleA}      & 110.41 & 377.13 \\
			Shi-Tomasi \cite{Shi1994GoodFT}       & 5.20 & 21.69 \\
			MSER \cite{Matas2004RobustWS}      & 64.19 & 221.79 \\
			KAZE \cite{Alcantarilla2012KAZEF}        & 105.85 & 298.43 \\
			AKZE \cite{Alcantarilla2013FastED}       & 18.02 & 56.93 \\
			FAST \cite{Rosten2010FasterAB}        & 0.89 & 1.88 \\
			LIFT \cite{Yi2016LIFTLI}        & 2209.03 & 4901.38 \\
			Key.Net \cite{Laguna2019KeyNetKD}    & 15.64 & 35.82 \\
			SuperPoint \cite{DeTone2018SuperPointSI} & 2.41 & 3.98 \\
			LF-Net \cite{Ono2018LFNetLL}     & 282.77 & 855.77 \\
			\midrule
			BALF (ours) & 8.15 & 29.02 \\
			\bottomrule
		\end{tabular}
	}
	\caption{\textbf{Computational cost (in millisecond)} for different keypoint detectors for a single image. It demonstrates that our network can run in real-time for a VGA resolution image, \ie 480$\times$640 pixels, which is commonly used for many robotic applications.}
	\label{table_efficiency}
	\vspace{-1.2em}
\end{table}

\section{Conclusion and Future Works}
\label{sec:conclusion}
We present the first pure MLP-based network for local feature detection. Our network takes advantages of the multi-axis gated MLP block and a squeeze-and-excitation MLP block to build a pure MLP-based image encoder. The detection module then apply differentiable channel-wise softmax operator for keypoint detection. 
Extensive evaluations have been conducted with both synthetic and real datasets. The experimental results demonstrate that our network delivers on-par detection performance on sharp images, and achieves superior performance with motion blurred images compared to prior works. 
Our network is also light-weight and is able to run in real-time for VGA resolution image, which further enables its application for time-constrained applications. 
It is usually required to build correspondences via feature matching for higher level applications, such as visual localization. We thus plan to design a network to learn motion blur robust descriptors for those detected keypoints as our future work.

{\small
\bibliographystyle{ieee_fullname}
\bibliography{egbib}

\begin{thebibliography}{10}\itemsep=-1pt

\bibitem{Aans2011InterestingIP}
Henrik Aan{\ae}s, A. Dahl, and Kim~Steenstrup Pedersen.
\newblock Interesting interest points.
\newblock {\em IJCV}, 97:18--35, 2011.

\bibitem{Alcantarilla2012KAZEF}
Pablo~Fern{\'a}ndez Alcantarilla, Adrien Bartoli, and Andrew~J. Davison.
\newblock Kaze features.
\newblock In {\em ECCV}, 2012.

\bibitem{Alcantarilla2013FastED}
Pablo~Fern{\'a}ndez Alcantarilla, Jes{\'u}s Nuevo, and Adrien Bartoli.
\newblock Fast explicit diffusion for accelerated features in nonlinear scale
  spaces.
\newblock In {\em BMVC}, 2013.

\bibitem{Bahat2017NonuniformBD}
Yuval Bahat, Netalee Efrat, and Michal Irani.
\newblock Non-uniform blind deblurring by reblurring.
\newblock In {\em ICCV}, pages 3306--3314, 2017.

\bibitem{Balntas2017HPatchesAB}
Vassileios Balntas, Karel Lenc, Andrea Vedaldi, and Krystian Mikolajczyk.
\newblock Hpatches: A benchmark and evaluation of handcrafted and learned local
  descriptors.
\newblock In {\em CVPR}, pages 3852--3861, 2017.

\bibitem{Bay2006SURFSU}
Herbert Bay, Tinne Tuytelaars, and Luc~Van Gool.
\newblock Surf: Speeded up robust features.
\newblock In {\em ECCV}, 2006.

\bibitem{Benbihi2019ELFEL}
Assia Benbihi, Matthieu Geist, and C{\'e}dric Pradalier.
\newblock Elf: Embedded localisation of features in pre-trained cnn.
\newblock In {\em ICCV}, pages 7939--7948, 2019.

\bibitem{Chen2022CycleMLPAM}
Shoufa Chen, Enze Xie, Chongjian Ge, Ding Liang, and Ping Luo.
\newblock Cyclemlp: A mlp-like architecture for dense prediction.
\newblock In {\em ICLR}, 2022.

\bibitem{Cho2009FastMD}
Sunghyun Cho and Seungyong Lee.
\newblock Fast motion deblurring.
\newblock In {\em ACM SIGGRAPH Asia}, 2009.

\bibitem{Choe2022PointMixerMF}
Jaesung Choe, Chunghyun Park, François Rameau, Jaesik Park, and In-So Kweon.
\newblock Pointmixer: Mlp-mixer for point cloud understanding.
\newblock In {\em ECCV}, 2022.

\bibitem{Csurka2018FromHT}
Gabriela Csurka and M. Humenberger.
\newblock From handcrafted to deep local invariant features.
\newblock {\em ArXiv}, abs/1807.10254, 2018.

\bibitem{DeTone2018SuperPointSI}
Daniel DeTone, Tomasz Malisiewicz, and Andrew Rabinovich.
\newblock Superpoint: Self-supervised interest point detection and description.
\newblock In {\em CVPRW}, pages 337--33712, 2018.

\bibitem{Dosovitskiy2021AnII}
Alexey Dosovitskiy, Lucas Beyer, Alexander Kolesnikov, Dirk Weissenborn,
  Xiaohua Zhai, Thomas Unterthiner, Mostafa Dehghani, Matthias Minderer, Georg
  Heigold, Sylvain Gelly, Jakob Uszkoreit, and Neil Houlsby.
\newblock An image is worth 16x16 words: Transformers for image recognition at
  scale.
\newblock In {\em ICLR}, 2021.

\bibitem{Dusmanu2019D2NetAT}
Mihai Dusmanu, Ignacio Rocco, Tom{\'a}s Pajdla, Marc Pollefeys, Josef Sivic,
  Akihiko Torii, and Torsten Sattler.
\newblock D2-net: A trainable cnn for joint description and detection of local
  features.
\newblock In {\em CVPR}, pages 8084--8093, 2019.

\bibitem{Gauglitz2011EvaluationOI}
Steffen Gauglitz, Tobias H{\"o}llerer, and Matthew~A. Turk.
\newblock Evaluation of interest point detectors and feature descriptors for
  visual tracking.
\newblock {\em IJCV}, 94:335--360, 2011.

\bibitem{Goldstein2012BlurKernelEF}
Amit Goldstein and Raanan Fattal.
\newblock Blur-kernel estimation from spectral irregularities.
\newblock In {\em ECCV}, 2012.

\bibitem{Gong2017FromMB}
Dong Gong, Jie Yang, Lingqiao Liu, Yanning Zhang, Ian~D. Reid, Chunhua Shen,
  Anton van~den Hengel, and Qinfeng Shi.
\newblock From motion blur to motion flow: A deep learning solution for
  removing heterogeneous motion blur.
\newblock In {\em CVPR}, pages 3806--3815, 2017.

\bibitem{Guo2022HireMLPVM}
Jianyuan Guo, Yehui Tang, Kai Han, Xinghao Chen, Han Wu, Chao Xu, Chang Xu, and
  Yunhe Wang.
\newblock Hire-mlp: Vision mlp via hierarchical rearrangement.
\newblock In {\em CVPR}, pages 816--826, 2022.

\bibitem{Hou2022VisionPA}
Qibin Hou, Zihang Jiang, Li Yuan, Mingg-Ming Cheng, Shuicheng Yan, and Jiashi
  Feng.
\newblock Vision permutator: A permutable mlp-like architecture for visual
  recognition.
\newblock {\em IEEE TPAMI}, 2022.

\bibitem{Hradi2015ConvolutionalNN}
Michal Hradis, Jan Kotera, Pavel Zemcik, and Filip Sroubek.
\newblock Convolutional neural networks for direct text deblurring.
\newblock In {\em BMVC}, 2015.

\bibitem{Hu2020SqueezeandExcitationN}
Jie Hu, Li Shen, Samuel Albanie, Gang Sun, and Enhua Wu.
\newblock Squeeze-and-excitation networks.
\newblock {\em IEEE TPAMI}, 42:2011--2023, 2020.

\bibitem{Krishnan2011BlindDU}
Dilip Krishnan, Terence Tay, and Rob Fergus.
\newblock Blind deconvolution using a normalized sparsity measure.
\newblock In {\em CVPR}, pages 233--240, 2011.

\bibitem{Kupyn2018DeblurGANBM}
Orest Kupyn, Volodymyr Budzan, Mykola Mykhailych, Dmytro Mishkin, and Jiri
  Matas.
\newblock Deblurgan: Blind motion deblurring using conditional adversarial
  networks.
\newblock In {\em CVPR}, pages 8183--8192, 2018.

\bibitem{Kupyn2019DeblurGANv2D}
Orest Kupyn, T. Martyniuk, Junru Wu, and Zhangyang Wang.
\newblock Deblurgan-v2: Deblurring (orders-of-magnitude) faster and better.
\newblock In {\em ICCV}, pages 8877--8886, 2019.

\bibitem{Laguna2019KeyNetKD}
Axel~Barroso Laguna, Edgar Riba, Daniel Ponsa, and Krystian Mikolajczyk.
\newblock Key.net: Keypoint detection by handcrafted and learned cnn filters.
\newblock In {\em ICCV}, pages 5835--5843, 2019.

\bibitem{Lenc2016LearningCF}
Karel Lenc and Andrea Vedaldi.
\newblock Learning covariant feature detectors.
\newblock In {\em ECCV Workshops}, 2016.

\bibitem{Leutenegger2011BRISKBR}
Stefan Leutenegger, Margarita Chli, and Roland~Y. Siegwart.
\newblock Brisk: Binary robust invariant scalable keypoints.
\newblock In {\em ICCV}, pages 2548--2555, 2011.

\bibitem{Li2018LearningAD}
Lerenhan Li, Jinshan Pan, Wei-Sheng Lai, Changxin Gao, Nong Sang, and
  Ming-Hsuan Yang.
\newblock Learning a discriminative prior for blind image deblurring.
\newblock In {\em CVPR}, pages 6616--6625, 2018.

\bibitem{Li2022BraininspiredMP}
Wenshuo Li, Hanting Chen, Jianyuan Guo, Ziyang Zhang, and Yunhe Wang.
\newblock Brain-inspired multilayer perceptron with spiking neurons.
\newblock In {\em CVPR}, pages 773--783, 2022.

\bibitem{Lian2022ASMLPAA}
Dongze Lian, Zehao Yu, Xing Sun, and Shenghua Gao.
\newblock As-mlp: An axial shifted mlp architecture for vision.
\newblock In {\em ICLR}, 2022.

\bibitem{Lin2014MicrosoftCC}
Tsung-Yi Lin, Michael Maire, Serge~J. Belongie, James Hays, Pietro Perona, Deva
  Ramanan, Piotr Doll{\'a}r, and C.~Lawrence Zitnick.
\newblock Microsoft coco: Common objects in context.
\newblock In {\em ECCV}, 2014.

\bibitem{Liu2021PayAT}
Hanxiao Liu, Zihang Dai, David~R. So, and Quoc~V. Le.
\newblock Pay attention to mlps.
\newblock In {\em NeurIPS}, 2021.

\bibitem{Liu2020SelfSupervisedLM}
Peidong Liu, Joel Janai, Marc Pollefeys, Torsten Sattler, and Andreas Geiger.
\newblock Self-supervised linear motion deblurring.
\newblock {\em IEEE Robotics and Automation Letters}, 5:2475--2482, 2020.

\bibitem{LoweDavid2004DistinctiveIF}
G LoweDavid.
\newblock Distinctive image features from scale-invariant keypoints.
\newblock {\em IJCV}, 2004.

\bibitem{Mansour2022ImagetoImageMF}
Youssef Mansour, Kang Lin, and Reinhard Heckel.
\newblock Image-to-image mlp-mixer for image reconstruction.
\newblock {\em ArXiv}, abs/2202.02018, 2022.

\bibitem{Matas2004RobustWS}
Jiri Matas, Ondřej Chum, Martin Urban, and Tom{\'a}s Pajdla.
\newblock Robust wide-baseline stereo from maximally stable extremal regions.
\newblock {\em Image Vis. Comput.}, 22:761--767, 2004.

\bibitem{Mikolajczyk2003APE}
Krystian Mikolajczyk and Cordelia Schmid.
\newblock A performance evaluation of local descriptors.
\newblock In {\em CVPR}, pages II--II, 2003.

\bibitem{Mikolajczyk2004ScaleA}
Krystian Mikolajczyk and Cordelia Schmid.
\newblock Scale \& affine invariant interest point detectors.
\newblock {\em IJCV}, 60:63--86, 2004.

\bibitem{Mikolajczyk2005ACO}
Krystian Mikolajczyk, Tinne Tuytelaars, Cordelia Schmid, Andrew Zisserman, Jiri
  Matas, Frederik Schaffalitzky, Timor Kadir, and Luc~Van Gool.
\newblock A comparison of affine region detectors.
\newblock {\em IJCV}, 65:43--72, 2005.

\bibitem{Nah2017DeepMC}
Seungjun Nah, Tae~Hyun Kim, and Kyoung~Mu Lee.
\newblock Deep multi-scale convolutional neural network for dynamic scene
  deblurring.
\newblock In {\em CVPR}, pages 257--265, 2017.

\bibitem{Ono2018LFNetLL}
Yuki Ono, Eduard Trulls, Pascal~V. Fua, and Kwang~Moo Yi.
\newblock Lf-net: Learning local features from images.
\newblock In {\em NeurIPS}, 2018.

\bibitem{Perrone2014TotalVB}
Daniel~J. Perrone and Paolo Favaro.
\newblock Total variation blind deconvolution: The devil is in the details.
\newblock In {\em CVPR}, pages 2909--2916, 2014.

\bibitem{Ren2018DeepND}
Wenqi Ren, Jiawei Zhang, Lin Ma, Jinshan Pan, Xiaochun Cao, Wangmeng Zuo, W.
  Liu, and Ming-Hsuan Yang.
\newblock Deep non-blind deconvolution via generalized low-rank approximation.
\newblock In {\em NeurIPS}, 2018.

\bibitem{Revaud2019R2D2RA}
J{\'e}r{\^o}me Revaud, Philippe Weinzaepfel, C{\'e}sar~Roberto de Souza, No'e
  Pion, Gabriela Csurka, Yohann Cabon, and M. Humenberger.
\newblock R2d2: Repeatable and reliable detector and descriptor.
\newblock In {\em NeurIPS}, 2019.

\bibitem{Rim2020RealWorldBD}
Jaesung Rim, Hoon~Sung Chwa, and Sunghyun Cho.
\newblock Real-world blur dataset for learning and benchmarking deblurring
  algorithms.
\newblock In {\em ECCV}, 2020.

\bibitem{Rosten2006MachineLF}
Edward Rosten and Tom Drummond.
\newblock Machine learning for high-speed corner detection.
\newblock In {\em ECCV}, 2006.

\bibitem{Rosten2010FasterAB}
Edward Rosten, Reid~B. Porter, and Tom Drummond.
\newblock Faster and better: A machine learning approach to corner detection.
\newblock {\em IEEE TPAMI}, 32:105--119, 2010.

\bibitem{Rublee2011ORBAE}
Ethan Rublee, Vincent Rabaud, Kurt Konolige, and Gary~R. Bradski.
\newblock Orb: An efficient alternative to sift or surf.
\newblock In {\em ICCV}, pages 2564--2571, 2011.

\bibitem{ILSVRC15}
Olga Russakovsky, Jia Deng, Hao Su, Jonathan Krause, Sanjeev Satheesh, Sean Ma,
  Zhiheng Huang, Andrej Karpathy, Aditya Khosla, Michael Bernstein,
  Alexander~C. Berg, and Li Fei-Fei.
\newblock {ImageNet Large Scale Visual Recognition Challenge}.
\newblock {\em IJCV}, 115(3):211--252, 2015.

\bibitem{Salahat2017RecentAI}
Ehab Salahat and Murad Qasaimeh.
\newblock Recent advances in features extraction and description algorithms: A
  comprehensive survey.
\newblock In {\em 2017 IEEE International Conference on Industrial Technology
  (ICIT)}, pages 1059--1063, 2017.

\bibitem{Savinov2017QuadNetworksUL}
Nikolay Savinov, Akihito Seki, Lubor Ladicky, Torsten Sattler, and Marc
  Pollefeys.
\newblock Quad-networks: Unsupervised learning to rank for interest point
  detection.
\newblock In {\em CVPR}, pages 3929--3937, 2017.

\bibitem{Schmid2004EvaluationOI}
Cordelia Schmid, Roger Mohr, and Christian Bauckhage.
\newblock Evaluation of interest point detectors.
\newblock {\em IJCV}, 37:151--172, 2004.

\bibitem{Schuler2016LearningTD}
Christian~J. Schuler, Michael Hirsch, Stefan Harmeling, and Bernhard
  Sch{\"o}lkopf.
\newblock Learning to deblur.
\newblock {\em IEEE TPAMI}, 38:1439--1451, 2016.

\bibitem{Pan2014DeblurringTI}
Jin shan Pan, Zhe Hu, Zhixun Su, and Ming-Hsuan Yang.
\newblock Deblurring text images via l0-regularized intensity and gradient
  prior.
\newblock In {\em CVPR}, pages 2901--2908, 2014.

\bibitem{Pan2016BlindID}
Jin shan Pan, Deqing Sun, Hanspeter Pfister, and Ming-Hsuan Yang.
\newblock Blind image deblurring using dark channel prior.
\newblock In {\em CVPR}, pages 1628--1636, 2016.

\bibitem{Shi1994GoodFT}
Jianbo Shi and Carlo Tomasi.
\newblock Good features to track.
\newblock In {\em CVPR}, pages 593--600, 1994.

\bibitem{Sun2015LearningAC}
Jian Sun, Wenfei Cao, Zongben Xu, and Jean Ponce.
\newblock Learning a convolutional neural network for non-uniform motion blur
  removal.
\newblock In {\em CVPR}, pages 769--777, 2015.

\bibitem{Suwanwimolkul2021LearningOL}
Suwichaya Suwanwimolkul, Satoshi Komorita, and Kazuyuki Tasaka.
\newblock Learning of low-level feature keypoints for accurate and robust
  detection.
\newblock pages 2261--2270, 2021.

\bibitem{Tang2022AnIP}
Yehui Tang, Kai Han, Jianyuan Guo, Chang Xu, Yanxi Li, Chao Xu, and Yunhe Wang.
\newblock An image patch is a wave: Phase-aware vision mlp.
\newblock In {\em CVPR}, pages 10925--10934, 2022.

\bibitem{Tao2018ScaleRecurrentNF}
Xin Tao, Hongyun Gao, Yi Wang, Xiaoyong Shen, Jue Wang, and Jiaya Jia.
\newblock Scale-recurrent network for deep image deblurring.
\newblock In {\em CVPR}, pages 8174--8182, 2018.

\bibitem{Tian2020D2DKE}
Yurun Tian, Vassileios Balntas, Tony Ng, Axel~Barroso Laguna, Y. Demiris, and
  Krystian Mikolajczyk.
\newblock D2d: Keypoint extraction with describe to detect approach.
\newblock In {\em ACCV}, 2020.

\bibitem{Tolstikhin2021MLPMixerAA}
Ilya~O. Tolstikhin, Neil Houlsby, Alexander Kolesnikov, Lucas Beyer, Xiaohua
  Zhai, Thomas Unterthiner, Jessica Yung, Daniel Keysers, Jakob Uszkoreit,
  Mario Lucic, and Alexey Dosovitskiy.
\newblock Mlp-mixer: An all-mlp architecture for vision.
\newblock In {\em NeurIPS}, 2021.

\bibitem{Touvron2022ResMLPFN}
Hugo Touvron, Piotr Bojanowski, Mathilde Caron, Matthieu Cord, Alaaeldin
  El-Nouby, Edouard Grave, Gautier Izacard, Armand Joulin, Gabriel Synnaeve,
  Jakob Verbeek, and Herv'e J'egou.
\newblock Resmlp: Feedforward networks for image classification with
  data-efficient training.
\newblock {\em IEEE TPAMI}, 2022.

\bibitem{Tu2022MAXIMMM}
Zhengzhong Tu, Hossein Talebi, Han Zhang, Feng Yang, Peyman Milanfar,
  Alan~Conrad Bovik, and Yinxiao Li.
\newblock Maxim: Multi-axis mlp for image processing.
\newblock In {\em CVPR}, pages 5759--5770, 2022.

\bibitem{Verdie2015TILDEAT}
Yannick Verdie, Kwang~Moo Yi, Pascal~V. Fua, and Vincent Lepetit.
\newblock Tilde: A temporally invariant learned detector.
\newblock In {\em CVPR}, pages 5279--5288, 2015.

\bibitem{Woo2018CBAMCB}
Sanghyun Woo, Jongchan Park, Joon-Young Lee, and In-So Kweon.
\newblock Cbam: Convolutional block attention module.
\newblock In {\em ECCV}, 2018.

\bibitem{Xiao2016LearningHF}
Lei Xiao, Jue Wang, Wolfgang Heidrich, and Michael Hirsch.
\newblock Learning high-order filters for efficient blind deconvolution of
  document photographs.
\newblock In {\em ECCV}, 2016.

\bibitem{Xie2021SegFormerSA}
Enze Xie, Wenhai Wang, Zhiding Yu, Anima Anandkumar, Jos{\'e}~Manuel
  {\'A}lvarez, and Ping Luo.
\newblock Segformer: Simple and efficient design for semantic segmentation with
  transformers.
\newblock 2021.

\bibitem{Xu2010TwoPhaseKE}
Li Xu and Jiaya Jia.
\newblock Two-phase kernel estimation for robust motion deblurring.
\newblock In {\em ECCV}, 2010.

\bibitem{Xu2014DeepCN}
Li Xu, Jimmy S.~J. Ren, Ce Liu, and Jiaya Jia.
\newblock Deep convolutional neural network for image deconvolution.
\newblock In {\em NeurIPS}, 2014.

\bibitem{Xu2013UnnaturalLS}
Li Xu, Shicheng Zheng, and Jiaya Jia.
\newblock Unnatural l0 sparse representation for natural image deblurring.
\newblock In {\em CVPR}, pages 1107--1114, 2013.

\bibitem{Xu2018MotionBK}
Xiangyu Xu, Jinshan Pan, Yujin Zhang, and Ming-Hsuan Yang.
\newblock Motion blur kernel estimation via deep learning.
\newblock {\em IEEE TIP}, 27:194--205, 2018.

\bibitem{Yan2016BlindIB}
Ruomei Yan and Ling Shao.
\newblock Blind image blur estimation via deep learning.
\newblock {\em IEEE TIP}, 25:1910--1921, 2016.

\bibitem{Yi2016LIFTLI}
Kwang~Moo Yi, Eduard Trulls, Vincent Lepetit, and Pascal~V. Fua.
\newblock Lift: Learned invariant feature transform.
\newblock In {\em ECCV}, 2016.

\bibitem{Yu2021S2MLPv2IS}
Tan Yu, Xu Li, Yunfeng Cai, Mingming Sun, and Ping Li.
\newblock S2-mlpv2: Improved spatial-shift mlp architecture for vision.
\newblock 2021.

\bibitem{Zamir2021MultiStagePI}
Syed~Waqas Zamir, Aditya Arora, Salman~Hameed Khan, Munawar Hayat,
  Fahad~Shahbaz Khan, Ming-Hsuan Yang, and Ling Shao.
\newblock Multi-stage progressive image restoration.
\newblock In {\em CVPR}, pages 14816--14826, 2021.

\bibitem{Zhang2018DynamicSD}
Jiawei Zhang, Jinshan Pan, Jimmy S.~J. Ren, Yibing Song, Linchao Bao, Rynson
  W.~H. Lau, and Ming-Hsuan Yang.
\newblock Dynamic scene deblurring using spatially variant recurrent neural
  networks.
\newblock In {\em CVPR}, pages 2521--2529, 2018.

\bibitem{Zhang2018LearningTD}
Linguang Zhang and Szymon~M. Rusinkiewicz.
\newblock Learning to detect features in texture images.
\newblock In {\em CVPR}, pages 6325--6333, 2018.

\bibitem{Zhang2017LearningDA}
Xu Zhang, Felix~X. Yu, Svebor Karaman, and Shih-Fu Chang.
\newblock Learning discriminative and transformation covariant local feature
  detectors.
\newblock In {\em CVPR}, pages 4923--4931, 2017.

\end{thebibliography}
}

\end{document}